\documentclass[letterpaper]{article} 
\usepackage[]{aaai25}  
\usepackage{times}  
\usepackage{helvet}  
\usepackage{courier}  
\usepackage[hyphens]{url}  
\usepackage{graphicx} 
\usepackage{natbib}  
\usepackage{caption} 
\usepackage{algorithm}
\usepackage{algorithmic}

\usepackage{newfloat}
\usepackage{listings}
\DeclareCaptionStyle{ruled}{labelfont=normalfont,labelsep=colon,strut=off} 
\floatstyle{ruled}
\newfloat{listing}{tb}{lst}{}
\floatname{listing}{Listing}
\usepackage{amsmath,amsfonts,bm}

\def\eqref#1{equation~\ref{#1}}

\def\1{\bm{1}}

\DeclareMathAlphabet{\mathsfit}{\encodingdefault}{\sfdefault}{m}{sl}
\SetMathAlphabet{\mathsfit}{bold}{\encodingdefault}{\sfdefault}{bx}{n}

\def\gA{{\mathcal{A}}}

\def\gD{{\mathcal{D}}}

\def\gN{{\mathcal{N}}}

\def\gP{{\mathcal{P}}}

\def\gU{{\mathcal{U}}}

\def\gX{{\mathcal{X}}}
\def\gY{{\mathcal{Y}}}

\DeclareMathOperator*{\argmin}{arg\,min}

\newcommand{\norm}[1]{\left\lVert#1\right\rVert}
\newcommand{\bcc}[1]{\left\{{#1}\right\}}

\newcommand{\bss}[1]{\left[{#1}\right]}
\newcommand{\ipp}[2]{\left\langle{#1},{#2}\right\rangle}
\newcommand{\abs}[1]{\left\vert#1\right\vert}

\usepackage{microtype}
\usepackage{graphicx}
\usepackage{subfigure}
\usepackage{booktabs} 

\usepackage{titletoc}

\usepackage{amsthm}
\theoremstyle{plain}

\theoremstyle{definition}

\theoremstyle{remark}

\usepackage{amsmath}
\usepackage{mathtools}

\usepackage{enumitem}

\usepackage{url}

\usepackage{algorithm,algorithmic}
\usepackage{multirow}

\usepackage{color-edits}
\addauthor{vc}{blue}

\usepackage{caption} 

\usepackage{array}
\newcolumntype{H}{>{\setbox0=\hbox\bgroup}c<{\egroup}@{}}

\newcommand*\samethanks[1][\value{footnote}]{\footnotemark[#1]}

\title{Learning Personalized Decision Support Policies}
\author {
    Umang Bhatt\textsuperscript{\rm 1,2}\thanks{Both authors contributed equally: order decided by a coin flip.},
    Valerie Chen\textsuperscript{\rm 3}\samethanks,
    Katherine M. Collins\textsuperscript{\rm 4},
    Parameswaran Kamalaruban\textsuperscript{\rm 2},
    Emma Kallina\textsuperscript{\rm 2,4},
    Adrian Weller\textsuperscript{\rm 2,4},
    Ameet Talwalkar\textsuperscript{\rm 3}
}
\affiliations {
    \textsuperscript{\rm 1}New York University\\
    \textsuperscript{\rm 2}The Alan Turing Institute\\
    \textsuperscript{\rm 3}Carnegie Mellon University\\
    \textsuperscript{\rm 4}University of Cambridge\\
    umangbhatt@nyu.edu, valeriechen@cmu.edu
}

\usepackage{bibentry}

\begin{document}

\maketitle

\begin{abstract}
Individual human decision-makers may benefit from different forms of support to improve decision outcomes, but when will each form of support yield better outcomes? In this work, we posit that personalizing access to decision support tools can be an effective mechanism for instantiating the appropriate use of AI assistance. Specifically, we propose the general problem of learning a \textit{decision support policy} that, for a given input, chooses which form of support to provide to decision-makers for whom we initially have no prior information. We develop \texttt{Modiste}, an interactive tool to learn personalized decision support policies. \texttt{Modiste} leverages stochastic contextual bandit techniques to personalize a decision support policy for each decision-maker. In our computational experiments, we characterize the expertise profiles of decision-makers for whom personalized policies will outperform offline policies, including population-wide baselines. Our experiments include realistic forms of support (e.g., expert consensus and predictions from a large language model) on vision and language tasks. Our human subject experiments add nuance to and bolster our computational experiments, demonstrating the practical utility of personalized policies when real users benefit from accessing support across tasks.
\end{abstract}

%

\section{Introduction} 

\begin{figure}[]
\centering
  \includegraphics[scale=0.35]{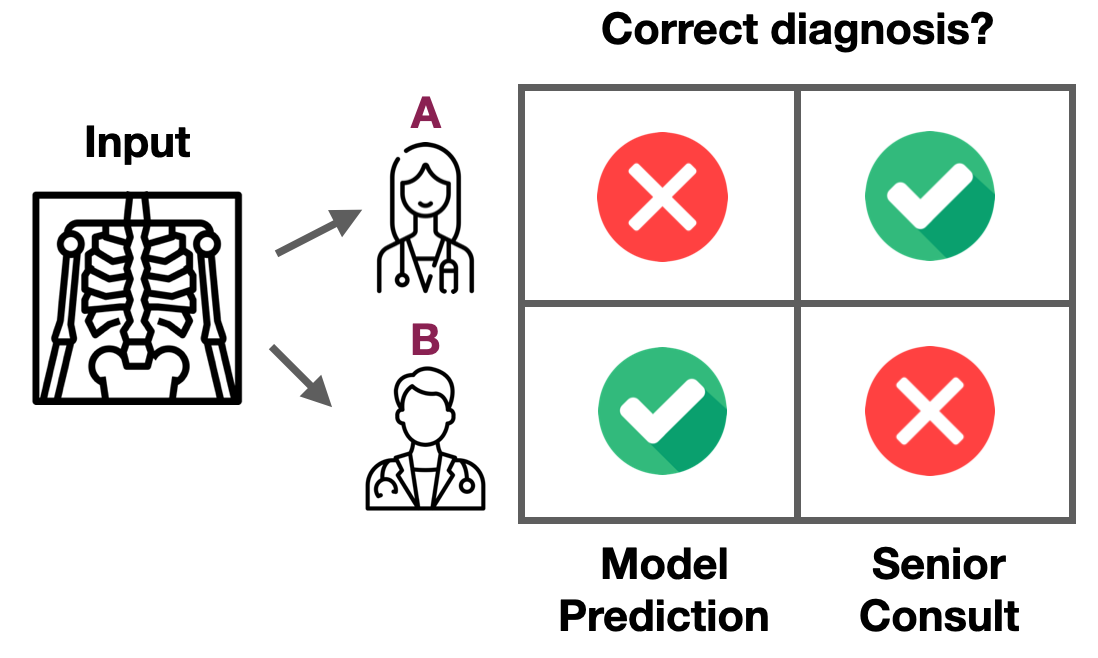}
  \caption{Depending on the input, decision-makers need different forms of decision support to make correct decisions. \texttt{Modiste} personalizes access to the right form of support at the right time for the right decision-maker online. Here, Alice would not benefit from model access, while Bob would not benefit from a senior consult.}
  \label{fig:motivation}
\end{figure}

Human decision-makers use various forms of support to inform their opinions before making a final decision~\citep{keen1980decision}.
Decision-makers with differing expertise may benefit from different forms of support on a given input~\citep{yu2024heterogeneity}.
For example, one radiologist may provide a better diagnosis of a chest X-ray by leveraging model predictions~\citep{kahn1994artificial} while another may perform better after viewing suggestions from senior radiologists~\citep{briggs2008role}: see Figure~\ref{fig:motivation}.
In this paper, we study how to improve decision outcomes by \emph{personalizing} which form of support we provide to a decision-maker on a case-by-case basis.

Since artificial intelligence (AI) is increasingly used as a form of decision support~\cite{lai2021towards}, even moving towards systems that could act as ``thought partners''~\cite{collins2024building}, responsible machine learning (ML) model deployment requires clarity on who should have access to model outputs and when model outputs can be safely exposed to decision-makers~\cite{amodei2016concrete}. 
Regulation increasingly calls for the ``effective and appropriate use'' of AI~\cite{EO14110}, requiring careful consideration of when models ought to be accessible to decision-makers.
In this paper, we formalize learning a \textit{decision support policy} that dictates for each individual decision-maker when additional support (e.g., LLM output) should be viewed and used for a given input.

While prior work has assumed access to offline human decisions under support~\citep{laidlaw2021uncertain,charusaie2022sample} or oracle queries of human behavior~\citep{de2018learning,mozannar2020consistent} to learn decision support policies, we argue that this data is unrealistic to obtain in practice across all available forms of support \emph{for a new decision-maker}.
Thus, for individuals for whom we have no prior information initially, we propose learning how to personalize support \emph{online}. 
We develop \texttt{Modiste},\footnote{While a ``modiste'' usually refers to someone who tailors clothing and makes dresses/hats, we use the term to capture our tool's ability to alter a policy to a decision-maker.} 
an interactive tool that leverages off-the-shelf stochastic contextual bandit algorithms ~\citep{li2010contextual} to learn decision support policies by modeling human prediction error under varying forms of support.


Our computational experiments explore the utility of personalization across multiple expertise profiles.
Based on these experiments, we characterize decision-maker expertise into profiles where personalized policies outperform offline ones, such as population-wide majority vote. We demonstrate that if there is no benefit of personalization, \texttt{Modiste} recovers the same performance as the best form of support. 

To validate \texttt{Modiste} on real users ($N=80$),  we conduct human subject experiments, where we explore forms of support that include expert consensus, outputs from an LLM, or predictions from a classification model.  
In contrast to prior work that only tests offline policies or evaluates in simulation, we demonstrate how \texttt{Modiste} can be used to learn personalized decision support policies online on both vision and language tasks. 
We emphasize our main contributions:



\textbf{1. Formalizing decision support policies.} We propose a formulation for learning a personalized decision support policy that selects the form of support that maximizes a given decision-maker's performance. We introduce \texttt{Modiste}, a tool to instantiate our formulation using existing methods from stochastic contextual bandits to model human prediction error under different forms of support.
We open-source \texttt{Modiste} as a tool to encourage the adoption of personalized decision support policies.

\textbf{2. Evaluating personalized policies in realistic settings.} We use \texttt{Modiste} to learn personalized policies for new decision-makers through both computational and human subject experiments on vision and language tasks. 
We characterize under which settings we would expect personalized policies to improve performance.
Our human subject experiments, where real users interact with \texttt{Modiste}, nuance our findings from the computational experiments on synthetic decision-makers, demonstrating the appropriate use of decision support has benefits in practice.

\section{Related Work}\label{sec:related}

\paragraph{Regulating AI Use.} 
Our study of decision support policies has implications for safely deploying ML models to interact with users. This topic is of heightened importance, particularly in light of recent calls for the ``effective and appropriate use'' of AI in US President Biden’s Executive Order~\cite{EO14110} and for clarity on when to ``decide not to use [an] AI system'' per the EU AI Act~\citep{EUAIAct}.
The disuse of AI can caution downstream misuse of models for assistive decision-making~\cite{brundage2018malicious}. The refusal to use AI assistance can be strategic to empower decision-makers, thus preventing their overreliance on models and encouraging their agency on the task at hand~\cite{gordon_meaningful_2020,barabas2022refusal}. 
Each decision-maker may require a different level of use to promote effective use of AI assistance in their decision-making~\cite{kirk2024prism}; for instance, experts and novices may prefer LLM access in different settings on theorem proving tasks~\cite{collins2023evaluating}.
Our experiments engage with such settings by learning when to provide LLM support for language-based tasks via a personalized policy. 

\paragraph{Decision Support.}
While various forms of decision support have been proposed, such as expert consensus~\citep{scheife2015consensus} and  changes to machine interfaces~\citep{roda2011human}, more recent forms of support focus on algorithmic tools where decision-makers are aided by 
machine learning (ML) models~\citep{phillips2012ai,gao2021human,bastani2022improving}. 
In some prior work, the human does not always make the final decision, such as those that learn to defer decisions from a model to a single decision-maker~\citep{madras2018predict,mozannar2020consistent} or others that jointly learn an allocation function between a model and a pool of decision makers~\citep{keswani2021towards,hemmer2022forming}.
In our setting, the \textit{human} is always the decision-maker, which includes settings where humans make the final decisions with support from ML models~\citep{green2019principles, lai2021towards}, as well as those where humans make decisions when provided with additional information beyond a model prediction, e.g., explanations~\citep{bansal2021does}, uncertainty~\citep{zhang2020effect}, 
conformal sets~\citep{babbar2022utility}.
While these studies \emph{always} show a single form of support, recent works consider adapting when AI support is shown to users with a fixed policy. 
\citet{ma2023should} fit a decision tree to offline user's decisions to decide when to show AI support to users, and 
\citet{buccinca2024towards} use offline reinforcement learning to estimate if AI support would be helpful, especially under time constraints~\cite{swaroop2024accuracy}.
Our work considers general forms of support, beyond when to show AI support, and formalizes \emph{learning} in which contexts each form of support should be provided to an unseen decision-maker online.
An extensive comparison to prior work is in the Appendix.

\paragraph{Prior Assumptions About Decision-Maker Information.} 
We briefly survey the assumptions made about the decision-maker when learning decision support policies. 
The model of the decision maker is either synthetic, thus lacking grounding in actual human behavioral data, or learned from a batch of \textit{offline} annotations~\citep{madras2018predict,okati2021differentiable, charusaie2022sample,gao2023learning}.
For a new decision-maker or a new form of support, this set of data would not be available in practice.
Instead, we propose to learn a decision support policy \emph{online} to circumvent these limitations. Few works use some aspect of online learning for different decision-making settings or under strict theoretical conditions, as we describe in the Appendix.



\begin{figure*}[tb]
\centering
  \includegraphics[scale=0.55]{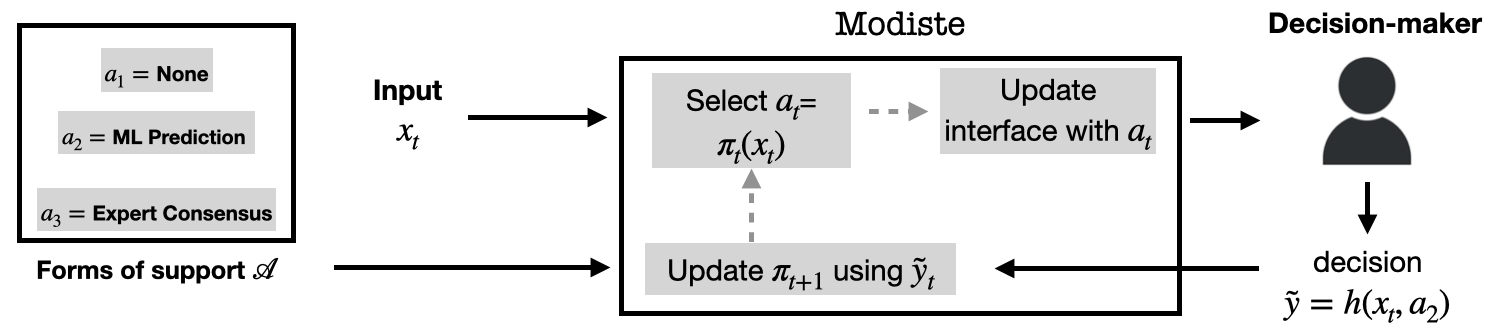}
  \caption{We illustrate the process of learning a decision support policy $\pi_t$ online to improve a decision-maker $h$'s performance. 
  Since assuming access to sufficient amounts of offline data is unreasonable in practice, our formulation learns a personalized policy \textit{online}; 
  each decision-maker's learned policy may differ from that of another decision-maker if they have decisions ($\tilde y$) and thus different expertise. 
  }
  \label{fig:dspoverview}
\end{figure*}

\section{Preliminaries}
We consider a human decision-making process with different forms of decision support. 
In our setting, decision-makers may be shown support when selecting an outcome from a fixed set of labels~\citep{seger2013categorization,lai2021towards}.

\paragraph{General Problem Formulation.}\label{sec:problem} 
Decision-makers perform a classification task in observation/feature space $\gX \subseteq \mathbb{R}^p$ and outcome/label space $\gY = \bss{K}$.
We operate in a stochastic setting where the data $(x,y) \in \gX \times \gY$ are drawn iid from a fixed, unknown data generating distribution $\gP$, an assumption that reflects typical decision-making settings~\citep{bastani2020online,bastani2022improving}. 
Importantly, we consider an action set $\gA$ corresponding to the forms of support available, which may consist of an individual piece of information (e.g., model prediction) or a particular combination of multiple pieces of information (e.g., model prediction and explanation). Given an observation $x \in \gX$, the human attempts to predict the corresponding label $y \in \gY$ using the support prescribed by an action $a \in \gA$. Note, we do not make assumptions on the specific forms of support $a$, but we provide multiple instantiations in our experiments.
The quality of predictions is measured using a 0-1 loss function, where $\ell(y, y') = 1$ for $y \ne y'$ and $\ell(y, y') = 0$ for $y = y'$.

\paragraph{Decision-Making Protocol.} 
A personalized decision support policy $\pi: \gX \to \Delta(\gA)$ outputs a form of support for a given input.
Let $\Pi$ denote the class of all stochastic decision support policies. Let $\gA = \bcc{\textsc{A}_1, \ldots, \textsc{A}_k}$, and $\pi(x)_{\textsc{A}_i}$ denote $\mathbb{P}[\textsc{A}_i \sim \pi(x)]$ for each $\textsc{A}_i \in \gA$. 
When the policy $\pi$ prescribes the support $\textsc{A}_i$, the human decision-maker makes the prediction $\widetilde{y}$ based on the observation $x$ and support $\textsc{A}_i$, i.e., the final prediction $\widetilde{y}$ is given by $\widetilde{y} = h(x, \textsc{A}_i)$. The human decision-making process with different forms of support is described below. For $t = 1,2,\dots,T$:
\begin{enumerate}[]
\item A data point $(x_t, y_t) \in \gX \times \gY$ is drawn iid from $\gP$.
\item A form of support $a_t \in \gA$ is selected using a decision support policy $\pi_t: \gX \to \Delta(\gA)$.
\item The human decision-maker makes the final prediction $\widetilde{y}_t = h(x_t, a_t)$ based on $x_t$ and $a_t$.
\item The human decision-maker incurs a loss $\ell(y_t, \widetilde{y}_t) = 1$ if $y_t \ne \widetilde{y}_t$ and $\ell(y_t, \widetilde{y}_t) = 0$ otherwise.
\end{enumerate}

\paragraph{Evaluation of $\pi$ via Expected Loss.}  The quality of a policy $\pi$ can be evaluated using the expected loss incurred by the decision-maker across the input space:
\begin{equation}\label{eq:risk}
    \begin{split}
        L_h(\pi) &~=~ \mathbb{E}_{(x,y) \sim \gP} \big[\mathbb{E}_{\textsc{A}_i \sim \pi(x)}[\ell (y, h(x, \textsc{A}_i))]\big] .
    \end{split}
\end{equation}
We distinguish this metric from the more standard notion of regret, which is typically used to analyze policies in an online learning setting~\citep{li2010contextual}; however, we cannot realize $\pi^*$ for an unseen decision-maker in practical scenarios. 
Thus, we rely on $L_h(\cdot)$ as a proxy metric for evaluating the effectiveness of $\pi$.

\section{\texttt{Modiste}: Learning Personalized Decision Support Policies}

We introduce \texttt{Modiste}, a tool to translate our problem formulation into an interactive interface for learning personalized policies. 
The workflow, outlined in Figure~\ref{fig:dspoverview}, comprises a learning component to update the personalized policy and an interface to customize the appropriate form of support for each input and each decision-maker.

\subsection{Learning Problem}
To model the decision-making process of a human decision-maker \emph{without access to their previous decisions}, we consider a stochastic contextual bandit set-up, where the forms of support are the arms, $\gX$ is the context space, and the policy $\pi$ can be learned online.
In Algorithm~\ref{alg:DSP}, we detail an algorithm for learning a policy $\pi^*$ that minimizes expected loss online. 
Our goal is to find an optimal decision support policy $\pi^*$ that minimizes $L_h(\pi)$. We can rewrite Eq.~\ref{eq:risk} as follows: 
$$L_h(\pi) = \mathbb{E}_x \big[\sum_{i=1}^k{\pi(x)_{\textsc{A}_i} \cdot r_{\textsc{A}_i}(x; h)}\big],$$ where $r_{\textsc{A}_i}(x; h) = \mathbb{E}_{y|x}[\ell (y, h(x, \textsc{A}_i))]$ is the human prediction error for input $x$ and support $\textsc{A}_i$. Then, it can be shown that the optimal policy takes the form $\pi^*(x) = \argmin_{\textsc{A}_i \in \mathcal{A}} r_{\textsc{A}_i}(x; h)$: see derivation in Appendix. 
For \texttt{Modiste} to run, we must maintain and update our estimate of human prediction error $r_{\textsc{A}_i} (x; h)$ (Step 1 in  Algorithm~\ref{alg:DSP}) and of our policy $\pi$ (Step 2 in  Algorithm~\ref{alg:DSP}).


To update the estimate of human prediction error (Step 1), \texttt{Modiste} implements two approaches to estimate $r_{\textsc{A}_i} (x; h)$ for all $x \in \gX$ and $\textsc{A}_i \in \gA$, but note that any online learning algorithm can be used to update the  $\gU_{r}$. 
We first consider \textbf{LinUCB}~\citep{li2010contextual}, a common online learning algorithm that approximates the expected loss $r_{\textsc{A}_i} (x; h)$ by a linear function $\widehat{r}_{\textsc{A}_i}(x; h) := \ipp{\theta_{\textsc{A}_i}}{x}$. Although the linearity assumption may not hold in general, we learn the parameters $\bcc{\theta_{\textsc{A}_i}: \textsc{A}_i \in \gA}$ using LinUCB with the instantaneous reward function $R(x, y, \textsc{A}_i; h) := - \ell(y, h(x, \textsc{A}_i))$. We then normalize the resulting $\widehat{r}_{\textsc{A}_i}(x; h)$ values to lie in the range $[0,1]$. 
The second algorithm we use is an intuitive $K$-nearest neighbor (\textbf{KNN}) approach, which is a simplified variant of KNN-UCB~\citep{guan2018nonparametric}. Here, we maintain an evolving data buffer $\gD_t$, which accumulates a history of interactions with the decision-maker. For any new observation $x$, we estimate $\widehat{r}_{\textsc{A}_i}(x; h)$ values by finding $K$-nearest neighbors in $\gD_t$ and computing the average error of these neighbors.

\begin{algorithm}
\caption{Learning a decision  support policy}\label{alg:rpu}
\begin{algorithmic}[1]
    \STATE \textbf{Input:} human decision-maker $h$
    \STATE \textbf{Initialization:} data buffer $\gD_0 = \{\}$; human error values $\{ \widehat{r}_{\textsc{A}_i, 0} (x; h) = 0.5 : x \in \gX, \textsc{A}_i \in \gA \}$; initial policy $\pi_1$
    \FOR{$t = 1, 2, \dots, T$}
    \STATE data point $(x_t, y_t) \in \gX \times \gY$ is drawn iid from $\gP$
    \STATE support $a_t \in \gA$ is selected using policy $\pi_t$
    \STATE human makes the prediction $\widetilde{y}_t$ based on $x_t$ and $a_t$
    \STATE human incurs the loss $\ell(y_t, \widetilde{y}_t)$
    \STATE update the buffer $\gD_t \gets \gD_{t-1} \cup \bcc{(x_t, a_t, \ell(y_t, \widetilde{y}_t))}$
    \STATE update the decision support policy: \vspace{-2mm}
    \begin{align*}
    \widehat{r}_{\textsc{A}_i, t} (x; h) ~\gets~& \gU_{r}(\widehat{r}_{\textsc{A}_i, t-1} (x; h), \gD_t), \quad \forall \textsc{A}_i \in \mathcal{A} \tag{Step 1} \\
    \pi_{t+1}(x) ~\gets~& \gU_{\pi}(\bcc{\widehat{r}_{\textsc{A}_i, t}}_i) \tag{Step 2}
    \end{align*}
    \ENDFOR{}
    \STATE \textbf{Output:} policy $\pi_\lambda^\text{alg} \gets \pi_{T+1}$
  \end{algorithmic}
  \label{alg:DSP}
\end{algorithm}

In practical settings where interactions are limited (like in our human subject experiments), the number of interactions $T$ tends to be relatively small, which renders pure exploratory policies infeasible~\citep{sutton2018reinforcement}. 
Thus in Step 2, we guide exploration of the policy via: $$\pi_{t+1}(x) = \argmin_{\textsc{A}_i \in \mathcal{A}} \widehat{r}_{\textsc{A}_i,t}(x; h) + b_{\textsc{A}_i,t}(x; h),$$ where $b_{\textsc{A}_i,t}(x; h)$ corresponds to some exploration bonus. In the Appendix, we provide implementations of \texttt{Modiste} with LinUCB and with KNN.

\subsection{\texttt{Modiste} Interface}

\begin{figure}[]
\centering
  \includegraphics[width=0.9\linewidth]{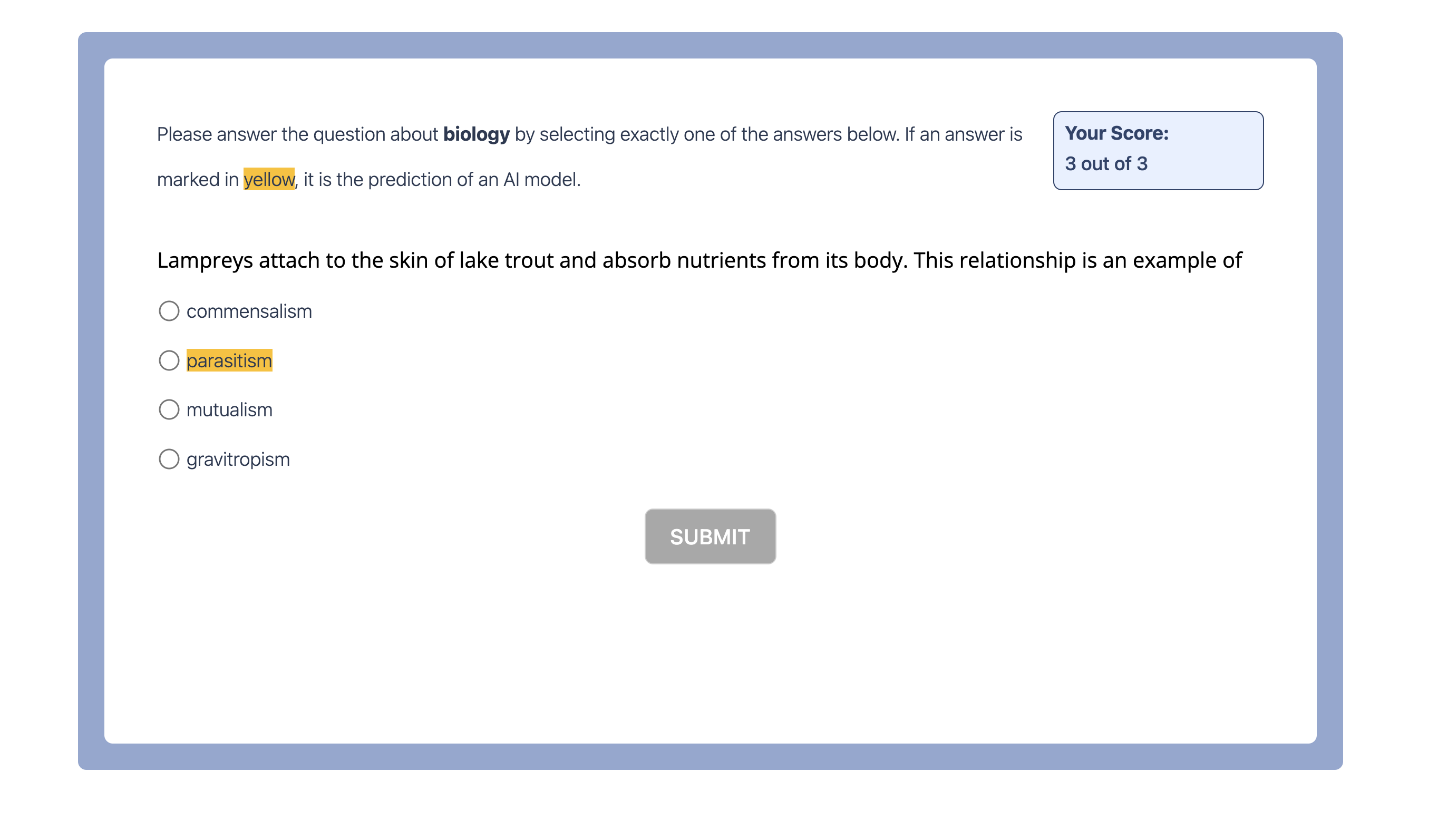}
  \caption{Example of the \texttt{Modiste} interface for MMLU-$2A$ where the human is provided responses from a LLM. 
  }
  \label{fig:hse-interface}
\end{figure}

We provide an extendable interface for the study and deployment of decision-support policies. At each time step, \texttt{Modiste} sends each user's predictions to a server running Algorithm~\ref{alg:DSP}, which identifies the next form of support for the next input. 
\texttt{Modiste} then updates the interface accordingly to reflect the selected form of support. 
Our tool can be flexiblely linked to crowdsourcing platforms like Prolific \citep{palan2018prolific}. 
We implement three common forms of support: (1) \textsc{Human Alone}, where the human makes the decision solely based on the input, (2) \textsc{Model Prediction}, which shows decision-makers a model's prediction for the given input~\citep{bastani2022improving}, and (3) \textsc{Expert Consensus}, which presents the user with a distribution over labels from multiple annotators~\citep{scheife2015consensus}.
In Figure~\ref{fig:hse-interface}, we provide an example screenshot of the interface. 
Participants are informed of their own correctness after each trial and the correctness of the form of support (e.g., model prediction) if support was provided, so that
participants can learn whether support ought to be relied upon.

\section{Experimental Set-up}

Before we evaluate \texttt{Modiste}, we overview the set-up of subsequent experiments, both computational and human subject. All other details are in the Appendix.


\subsection{Decision-making Tasks}\label{sec:tasks}

Following prior studies on human-AI interactions~\citep{babbar2022utility,NEURIPS2023_61355b9c,lee2023evaluating}, the decision-making tasks in our experiments center around the following vision and language datasets:
\begin{enumerate}
    \item \emph{CIFAR}-10~\citep{krizhevsky2009learning}, a 10-class image classification dataset;
    \item  \emph{MMLU}~\citep{hendrycks2020measuring}, a multi-task text-based benchmark that tests for knowledge and problem-solving ability across $57$ topics in both the humanities and STEM.
\end{enumerate}

In terms of the size of $|\mathcal{A}|$, we let $kA$ denote when there are $k$ forms of support for a task. We focus on $k=2$ or $3$, which captures a buffet of real-world scenarios in prior work where decision-makers have a few tools at their disposal.
In particular, the two action setting covers practical use cases where a decision-maker has the option of using a model or not. The learned decision support policy then reflects appropriate use, as the model would be hidden when a decision-maker does not need it. 
While the forms of support we consider are common in practice~\citep{lai2021towards}, our choices of support are not intended to exhaustively demonstrate the diverse forms of support that \texttt{Modiste} can handle.
We now describe our two main tasks, which are designed to be accessible to crowdworkers and will be featured in both the computational and human subject experiments.\footnote{In the Appendix, we include computational experiments for two additional tasks (Synthetic-$2A$ and CIFAR-$2A$), and experiments where we vary the size of $k$.} 

\paragraph{CIFAR-$3A$.} In this task, we consider three forms of support: \textsc{Human Alone}, \textsc{Model}, \textsc{Consensus}.
Our goal is to construct a setup reflecting a realistic setting in which different forms of support result in different strengths and weaknesses for decision-makers. 
To instantiate this setting, we deliberately corrupt images of different classes to evoke performance differences -- necessitating that a decision-maker appropriately calibrate when to rely on each form of support. 
We consider $5$ of the animal classes in CIFAR-10; of these, we never corrupt images of Birds, do not corrupt images of Deers and Cats for the \textsc{Model}, and do not corrupt images of Horses and Frogs for the \textsc{Consensus}.

\paragraph{MMLU-$2A$.} The two forms of support are \textsc{Human Alone} and \textsc{LLM}, where the human is provided responses generated from InstructGPT3.5, \texttt{text-davinci-003},~\citep{instructgpt} using the same few-shot prompting scheme for MMLU as~\citet{hendrycks2020measuring}. 
We conducted pilot studies to select a subset of topics where the accuracy of the LLM and average human accuracy vary. 
We choose the following topics: Computer Science, US Foreign Policy, High School Biology, and Elementary Mathematics.
The goal of this task is to evaluate whether we can learn personalized support \emph{``in-the-wild,''} where we naturally expect people to excel at different topics,
akin to real-world settings where decision-makers may have varying expertise. 

\subsection{Baselines and Other Parameters} 

\noindent \textbf{Algorithms and Baselines.} We compare personalized policies, learned using Algorithm~\ref{alg:DSP} with LinUCB and with KNN reporting results as \texttt{Modiste}-LinUCB and \texttt{Modiste}-KNN respectively, against the following offline policies:  
\begin{itemize}[leftmargin=*]
    \item \emph{Human + Support}, where the decision-maker \emph{always} receives the same form of support: $\pi(x) = A_i$ for all $x$. In CIFAR-$3A$, there are 3 fixed support baselines, corresponding to each form of support. In MMLU-$2A$, there are 2 fixed support baselines.
    \item \emph{Population-wide,} where the decision-maker receives a form of support based on the majority vote from 10 learned policies (breaking ties at random).
    For this baseline, the form of support may vary across contexts but is not personalized to individual needs. This baseline is akin to recent offline policy learning~\cite{ma2023should,buccinca2024towards}.
\end{itemize}

\noindent \textbf{Number of Interactions.} 
While more interactions (higher $T$) provide more data points to estimate each $r_{\textsc{A}_i}$, we need to consider what a realistic value of $T$ is given constraints of working with real humans (e.g., limited attention and cognitive load). 
In online learning, $T$ is usually unreasonably large, on the order of thousands~\cite{li2010contextual,guan2018nonparametric}.
Via pilot studies, we found that 100 CIFAR images or 60 MMLU questions were a reasonable number of decisions to make within 20-40 minutes (a typical time limit for an online study), which we use throughout our experiments.


\begin{table*}[htb!]
\centering
\footnotesize
\resizebox{\textwidth}{!}{\begin{tabular}{cccc}
\toprule
\multicolumn{1}{c}{Algorithm}        & \multicolumn{1}{c}{Invariant} & \multicolumn{1}{c}{Strictly Better} & Varying \\ \midrule
\multicolumn{1}{c}{\textsc{H-Only}}           
& \multicolumn{1}{c}{$0.00 \pm 0.01$}        
& \multicolumn{1}{c}{$0.09 \pm 0.08$}       
&   $0.50 \pm 0.06$              \\ 
\multicolumn{1}{c}{\textsc{H-Model}}     
& \multicolumn{1}{c}{$0.00 \pm 0.01$}        
& \multicolumn{1}{c}{$0.22 \pm 0.19$}        
&   $0.35 \pm 0.05$              \\ 
\multicolumn{1}{c}{\textsc{H-Consensus}} 
& \multicolumn{1}{c}{$0.00 \pm 0.01$}        
& \multicolumn{1}{c}{$0.23 \pm 0.13$}        
&    $0.27 \pm 0.08$             \\ 
\multicolumn{1}{c}{Population}      
& \multicolumn{1}{c}{$0.00 \pm 0.02$}        
& \multicolumn{1}{c}{$0.18 \pm 0.08$}        
&    $0.15 \pm 0.03$             \\ 
\multicolumn{1}{c}{\texttt{Modiste}-LinUCB}           
& \multicolumn{1}{c}{$0.00 \pm 0.01$}        
& \multicolumn{1}{c}{$0.17 \pm 0.05$}        
&    $0.19 \pm 0.05$             \\ 
\multicolumn{1}{c}{\texttt{Modiste}-KNN}              
& \multicolumn{1}{c}{$0.00 \pm 0.01$}        
& \multicolumn{1}{c}{$0.06 \pm 0.01$}       
&    $\mathbf{0.08 \pm 0.02}$             \\ \bottomrule
\end{tabular}
\quad
\begin{tabular}{cccc}
\toprule
\multicolumn{1}{c}{Algorithm}  & \multicolumn{1}{c}{Invariant}         & \multicolumn{1}{c}{Strictly Better} & Varying         \\ \midrule
\multicolumn{1}{c}{\textsc{H-Only}}     
& \multicolumn{1}{c}{$0.01 \pm 0.01$} 
& \multicolumn{1}{c}{$0.18 \pm 0.17$} 
& $0.22 \pm 0.12$  \\
\multicolumn{1}{c}{\textsc{H-LLM}} 
& \multicolumn{1}{c}{$0.01 \pm 0.01$} 
& \multicolumn{1}{c}{$0.18 \pm 0.21$} 
& $0.12 \pm 0.17$ \\ 
\multicolumn{1}{c}{Population} 
& \multicolumn{1}{c}{$0.00 \pm 0.02$} 
& \multicolumn{1}{c}{$0.19 \pm 0.07$} 
& $0.12 \pm 0.09$ \\ 
\multicolumn{1}{c}{\texttt{Modiste}-LinUCB}     
& \multicolumn{1}{c}{$0.00 \pm 0.01$} 
& \multicolumn{1}{c}{$0.12 \pm 0.03$} 
& $0.07 \pm 0.04$ \\ 
\multicolumn{1}{c}{\texttt{Modiste}-KNN}        
& \multicolumn{1}{c}{$0.01 \pm 0.01$} 
& \multicolumn{1}{c}{$0.05 \pm 0.03$} 
& $\mathbf{0.05 \pm 0.03}$ \\\bottomrule
\end{tabular}}
\caption{We evaluate \texttt{Modiste} across three expertise profiles. 
We compute the average excess loss $L_h(\pi)-L_h^{opt}$ (lower is better), and standard deviation across individuals in each expertise profile for both CIFAR-$3A$ (Left) and MMLU-$2A$ (Right).
$L_h(\pi)$ is computed by averaging across the last 10 steps of 100 total time steps. We \textbf{bold} the variant with the lowest excess loss that is statistically significant from the other variants. Note that this only occurs in the ``varying'' expertise setting.}
\label{tab:generalsetting}
\vspace{-.25cm}
\end{table*}

\section{Computational Evaluation}\label{sec:comp_exp}

Decision-makers may have different ``expertise'' (i.e., strengths and weaknesses) across input space $\mathcal{X}$ under each form of support. 
We evaluate \texttt{Modiste} using simulated human behavior to capture diverse decision-maker expertise. 

\subsection{Expertise Profiles}


We can capture an individual $h$'s expertise via an \emph{expertise profile}, which is defined over the input space $\gX$. We divide $\gX$ into disjoint regions (i.e., $\gX = \cup_{j \in [N]} \gX_j$); these regions could be defined by class labels or by covariates, depending on the task.\footnote{While we instantiate decision-makers this way, \texttt{Modiste} does not take the expertise profiles or how they were constructed (e.g., the regions) as input.}
We let $r_{\textsc{A}_i}(\gX_j; h)$ denote $h$'s average prediction error under support $\textsc{A}_i$ across region $\gX_j$.

\paragraph{Human-informed synthetic decision-makers.} 
To construct expertise profiles with \emph{realistic} values for each $r_{\textsc{A}_i}(\gX_j; h)$, we collect data on user decisions across different users and then calculate individual $r_{\textsc{A}_i}(\gX_j; h)$. 
The set of participant expertise profiles form a population of decision-makers that we refer to as \textit{human-informed synthetic decision-makers}. From the estimated $r_{\textsc{A}_i}(\gX_j; h)$ of each human-informed synthetic decision-maker, we can simulate decision-maker behavior.

To construct human-informed synthetic decision-makers, we recruited 20 participants from Prolific (10 for CIFAR-$3A$ and 10 for  MMLU-$2A$).
We use the same recruitment scheme as the larger human subject experiment described in the Appendix.
We define regions of expertise over class labels for CIFAR-$3A$ and over question topics for MMLU-$2A$, as we expect $r_{\textsc{A}_i}(x; h)$ to be roughly constant for $x \in \gX_j$ where $\gX_j$ is defined by a class label or question topic.
We showed each participant similar inputs with different forms of support to estimate $r_{\textsc{A}_i}(\gX_j; h)$ for each support $\textsc{A}_i$ in each region $\gX_j$. 
On each trial, each participant is randomly assigned a form of support; trials are approximately balanced by the type of support and grouping (i.e., topic or class). 
We compute participant accuracy averaged over all trials: 100 for CIFAR-$3A$, 60 for MMLU-$2A$. 
We denote expertise profiles as follows: if there were three regions in the input space, an individual's expertise profile under support $\textsc{A}_i$ would be written as $r_{\textsc{A}_i} = [0.7, 0.1, 0.7]$, meaning the individual incurs a loss of $0.7$ on $\gX_1$, $0.1$ on $\gX_2$, and $0.7$ on $\gX_3$.

\paragraph{Policies for each profile.}
Based our pilot study, we define the three expertise profiles and what kind of decision support policy we expect to be learned for each: 

\begin{itemize}[leftmargin=*]
    \item \textbf{Approximately Invariant} expertise across all the regions under different forms of support, i.e., $r_{\textsc{A}_1}(\gX_j; h) \approx r_{\textsc{A}_2}(\gX_j; h) \approx \cdots \approx r_{\textsc{A}_k}(\gX_j; h), \forall j \in [N]$. Both a random decision support policy and a policy that selects a fixed form of support would suffice for such a profile.
    \item \textbf{Varying} expertise where a decision-maker excels in some areas but benefits from support in areas beyond their training~\citep{schvaneveldt1985measuring}, i.e., $r_{\textsc{A}_1}(\gX_j; h) \leq r_{\textsc{A}_2}(\gX_j; h)\text{ and } r_{\textsc{A}_2}(\gX_k; h) \leq r_{\textsc{A}_1}(\gX_k; h), \text{for some } j,k \in [N]$. For this expertise profile, we expect the decision support policy to select different forms of support in different regions. 
    The quantity of $|r_{\textsc{A}_1}$ - $r_{\textsc{A}_2}|$ for a region will dictate how efficiently the policy can be learned. 
    \item \textbf{Strictly Better} expertise 
    (e.g., $\textsc{A}_1 \succ \textsc{A}_2 \succ \cdots \succ \textsc{A}_k$) 
    that is uniformly maintained across all the regions, i.e., $r_{\textsc{A}_1}(\gX_j; h) \leq r_{\textsc{A}_2}(\gX_j; h) \leq \cdots \leq r_{\textsc{A}_k}(\gX_j; h), \forall j \in [N]$. A decision support policy should learn the fixed form of support to use for all inputs.

\end{itemize}
Per the task design, we find participants generally only display varying expertise profiles on CIFAR-$3A$ while we find instances of all three expertise profiles on MMLU-$2A$.

\subsection{When is Personalization Useful?}


We investigate how personalized policies compare against offline baselines under each expertise profile (Table~\ref{tab:generalsetting}). 
We verify that learning decision support policies are not helpful for decision-makers with ``invariant'' expertise profiles. 
For individuals who fall under the ``varying'' profiles, we find at least one personalized policy outperforms offline policies and learns a policy that is significantly closer to the decision-maker's optimal performance.
This is because a personalized policy identifies \textit{which} form of support is better in each context, compared to fixed offline policies which \textit{always} show one form of support or to the population-wide variant, which may not provide the correct form of support to each individual. 
For individuals in the ``strictly better'' profile, while we do not find a statistically significant difference from the baselines due to the large variance in fixed policies (e.g., they work well for some decision-makers but poorly for others), we observe that much smaller variance with \texttt{Modiste}, particularly using KNN.
For most individuals, the population-wide baseline performs poorly, emphasizing the need for personalization of decision support.
Misalignment between the population policy and the optimal policy of the new decision-maker demonstrably leads to an ineffective use of decision support.
We note that KNN generally outperforms LinUCB, the latter of which can be saddled by its implicit linearity assumption. 
We further study the effect of various parameters, e.g., 
exploration parameters, KNN parameters, embedding size, and the number of interactions in the Appendix.

\section{\texttt{Modiste} with Real Users}\label{sec:user_study}


To validate whether \texttt{Modiste} can improve decision-maker performance in practice, we run a series of human subject experiments (i.e.,  ethics-reviewed studies with real human participants). 
We first introduce the set-up of our user study; additional information can be found in the Appendix.


\paragraph{Recruitment.} We recruit a total of $80$ crowdsourced participants from Prolific to interact with \texttt{Modiste} ($N=30$ and $N=50$ for CIFAR-$3A$ and MMLU-$2A$, respectively). We recruit more participants for MMLU-$2A$, as we expect greater individual differences in regions where support is needed, e.g., some participants may be good at mathematics and struggle in biology, whereas others may excel in biology.

Each participant is assigned to only one task. 
Within a task, participants are randomly assigned to one algorithm variant; an equal number of participants are included per variant (i.e., 10 for MMLU and 5 for CIFAR).
Participants are required to reside in the United States and speak English as a first language. Participants are paid at a base rate of \$9/hr, and are told they may be paid an optional bonus up to \$10/hr based on the number of correct responses. We allot 25-30 minutes for the CIFAR task and 30-40 for MMLU, as each MMLU question takes more effort. We applied the bonus to all participants in all studies.
We run an ANOVA with Tukey HSD across the conditions for each task.

\begin{figure*}[htb]
    \centering
    \begin{minipage}{0.47\textwidth}
        \centering
        \includegraphics[width=\textwidth]{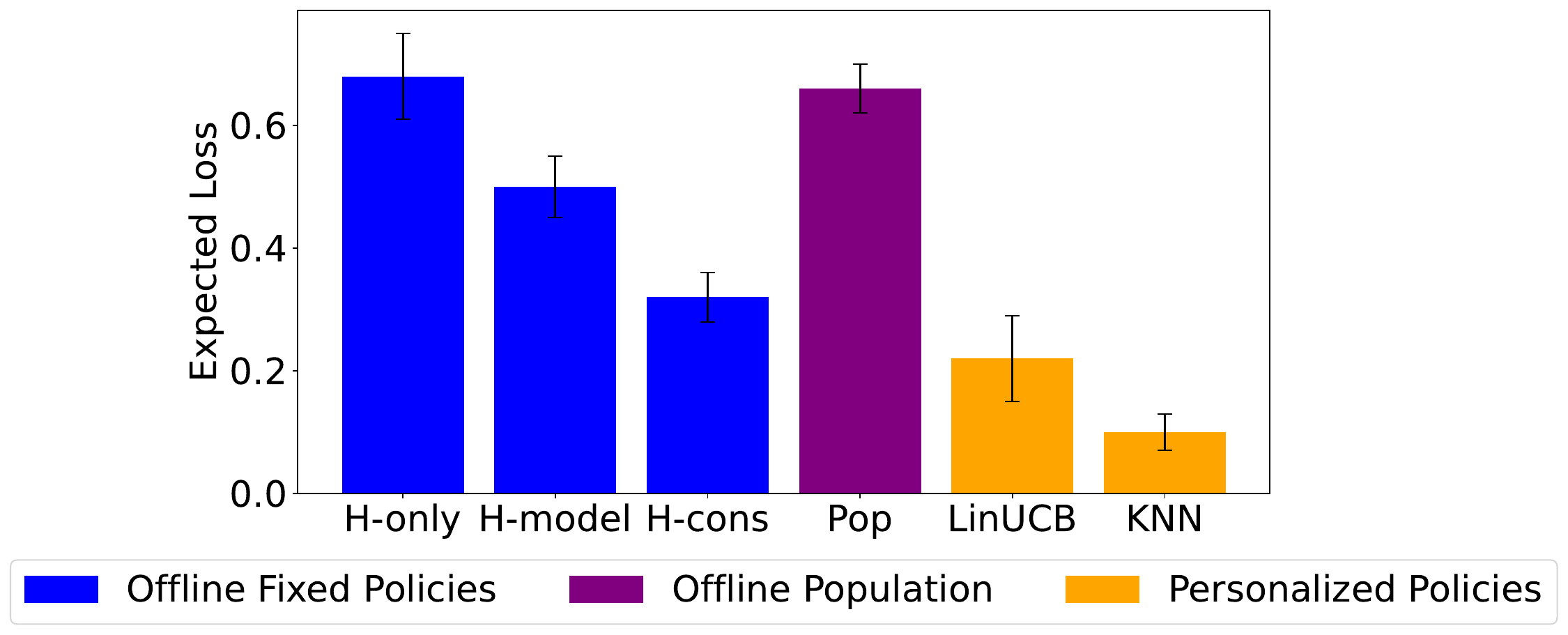} 
    \end{minipage}
    \hfill
    \begin{minipage}{0.45\textwidth}
        \centering
        \includegraphics[width=\textwidth]{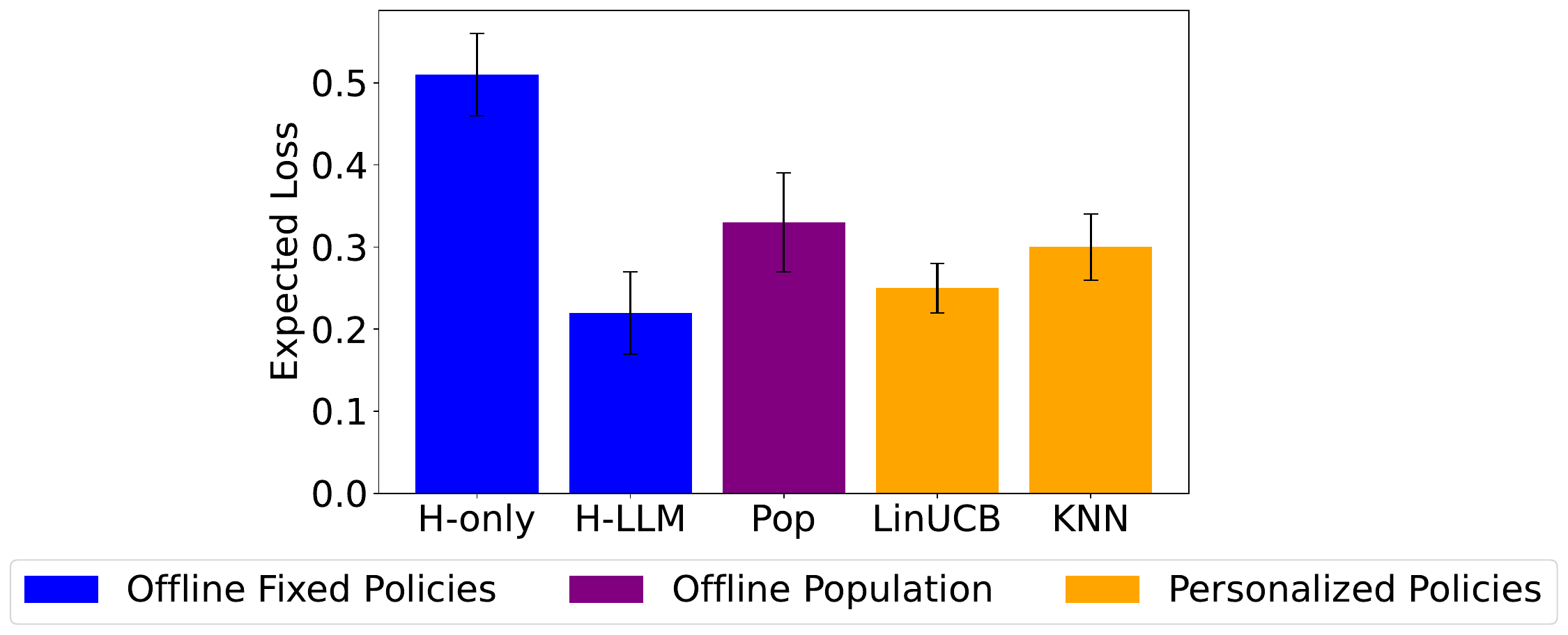} 
    \end{minipage}
    \caption{We report expected average loss $L_h(\pi)$ (lower is better) and standard error in the last 10 trials by Prolific participants for each algorithm, with CIFAR conditions on the left and MMLU conditions on the right. In the CIFAR setting, where individuals typically exhibit ``varying'' expertise profiles, we see significant benefits from using \texttt{Modiste}, particularly in the KNN setting. 
    While we observe that most individuals in the MMLU condition exhibit ``strictly better'' expertise, which means personalized policies typically only perform as well as the best baseline, we still observe instances of deferred decisions to the human on a case-by-case basis---see Figure~\ref{fig:hse-mmlu-topology}.
    }
    \label{fig:barplot_results}
\end{figure*}

\paragraph{\texttt{Modiste} outperforms baselines for ``varying'' expertise profiles.}
By design, the CIFAR-$3A$ task compels ``varying'' profiles: \texttt{Modiste}'s forte. 
This is reflected quantitatively in Figure~\ref{fig:barplot_results} (left), where both \texttt{Modiste} variants have lower expected losses than any of the offline policies. 
In particular, KNN achieves statistically significantly lower expected loss over all variants ($p<0.001$ across all pairs), including the population-wide policy learned from the pilot study. When visualizing the learned decision support policies, we observe that \texttt{Modiste} can reconstruct near-optimal policies, as depicted in the Appendix. 



\begin{figure*}[htb]
    \centering
    \begin{minipage}{0.98\textwidth}
        \centering
        \includegraphics[width=\textwidth]{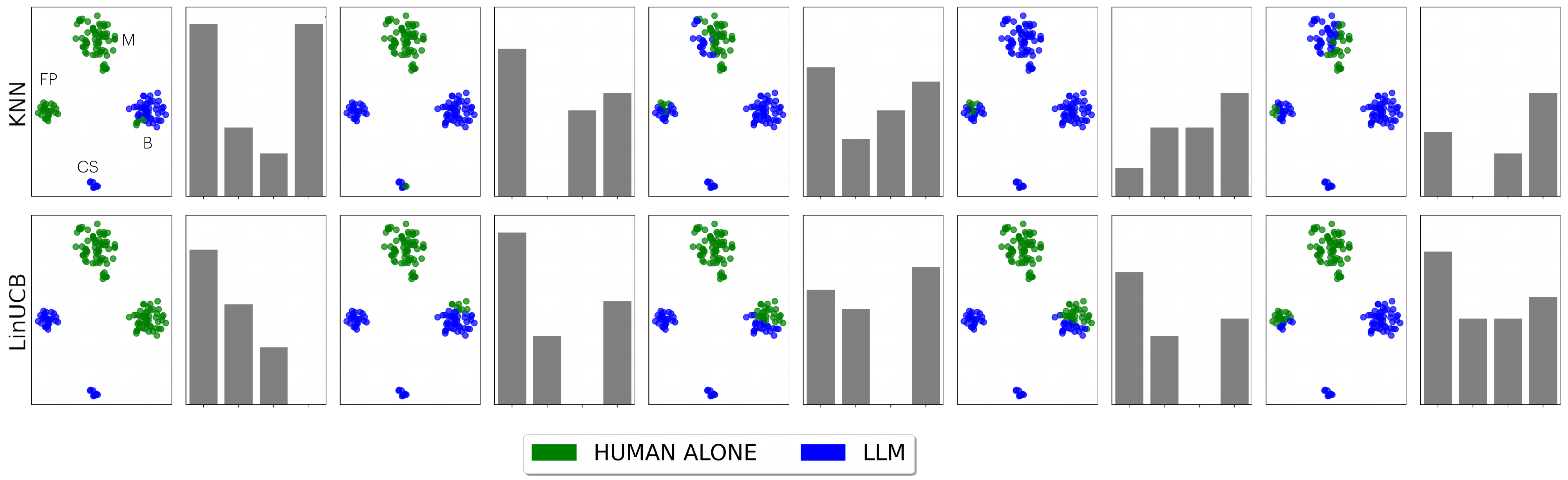} 
    \end{minipage}
    \caption{
    Snapshots of the learned decision support policies computed at the end of the study for 10 participants on the MMLU task. 
    The forms of support are colored in t-SNE embedding space.  
    All participants exhibit distinct policies across input space. 
    The bar plot to the right of each scatter plot shows the relative performance of that decision-maker \textit{alone} in each category, ordered from left to right as M=Mathematics, B=Biology, CS=Computer Science, FP=Foreign Policy per subplot.
    When a decision-maker performs well alone, \texttt{Modiste} learns policies to empower that decision-maker without LLM access. 
    For example, the individual in the top left is highly competent at both Mathematics and Foreign Policy; the learned decision support policy reflects this. 
    }
        \label{fig:hse-mmlu-topology}
\end{figure*}


\paragraph{\texttt{Modiste} matches the best baseline for ``strictly better'' profiles.} 
Polymaths are rare; we observe no different in our human subject experiments on MMLU, as most participants are better with LLM access, which places them in the ``strictly better'' category. 
While participants with both \texttt{Modiste} variants outperform the \textsc{H-Only} baseline ($p<0.01$), we observe that on average \texttt{Modiste} settings are no different than \textsc{H-LLM}, the condition where people always have access to the LLM, as shown in Figure~\ref{fig:barplot_results} (right).
This confirms our hypothesis from the computational experiments.
Further, while the average performance of \texttt{Modiste} variants is similar to that of the offline fixed or population-wide policies in Figure~\ref{fig:barplot_results} (right), the variance is significantly smaller---particularly with KNN.

\paragraph{\texttt{Modiste} can facilitate appropriate use of decision support.} 
Since the LLM only excels at three of the four topics on the MMLU task, \texttt{Modiste} learns in many cases to defer to human judgment on Mathematics questions, particularly when the human has strong expertise.
In Figure~\ref{fig:hse-mmlu-topology}, we visualize the learned decision support policies of various individuals in the study and illustrate how \texttt{Modiste}  yields policies that provide support on different topics for different decision-makers. 

\paragraph{Limitations.} 
In this work, we consider the classification setting where we get immediate feedback (e.g., we can calculate the loss to update $\pi$). Future work can consider more complex decision-making tasks that may require extending to a delayed feedback setting or to a different cognitive task (e.g., planning or perception). 
Though \texttt{Modiste} is promising, we note that significant issues can arise when decision-makers blindly rely on decision support~\citep{buccinca2020proxy,chen2023understanding}, especially when the support is erroneous or ineffective; such over-reliance requires careful attention to prevent. 
Further, our problem definition hinges on domain experts defining the available forms of support (i.e., we need a clearly defined $\mathcal{A}$ to use \texttt{Modiste}). 
In practice, this may prove difficult, as one may not know how to define specific forms of support or decision-makers may have access to varying support sets.

\section{Conclusion}
A decision support policy captures when and which form of support should be provided to improve a decision-maker's performance. 
The selective use of AI-based decision support helps instantiate the ``appropriate use'' clauses in emerging regulation~\cite{EO14110}, as we only provide AI assistance to decision-makers as and when it is beneficial to them.
We introduce \texttt{Modiste}, an interactive tool for learning a decision support policy for each decision-maker using contextual bandits. 
To the best of our knowledge, we are the first to learn and validate such a policy online for unseen decision-makers.
Within our \texttt{Modiste} interface, we instantiate two variants of Algorithm~\ref{alg:DSP} using existing stochastic contextual bandit tools, namely LinUCB and online KNN.
Our computational and human subject experiments highlight the importance---and feasibility---of personalizing decision support policies for individual decision-makers.
Our human subject experiments show promise, as we personalize decision support policies in few iterations yet find nuances in decision-makers' need for support: some unskilled decision-makers uniformly benefit from LLM access, while others only need LLMs for some tasks.
While encouraging rich cross-talk between domain experts and practitioners, future work integrating \texttt{Modiste} into existing decision-making workflows would pave a route towards responsible use of AI as decision support.

\section*{Acknowledgements}
UB acknowledges support from DeepMind and the Leverhulme Trust via the Leverhulme Centre for the Future of Intelligence (CFI). AW acknowledges support from a Turing AI Fellowship under grant EP/V025279/1, EPSRC grants EP/V056522/1 and EP/V056883/1, and the Leverhulme Trust via CFI. KMC acknowledges support from a Marshall Scholarship and the Cambridge Trust. PK acknowledges support from The Alan Turing Institute. AT acknowledges support from the National Science Foundation grants IIS1705121, IIS1838017, IIS2046613, IIS2112471, an Amazon Web Services Award, a Facebook Faculty Research Award, funding from Booz Allen Hamilton Inc., and a Block Center Grant. 
This work is supported (in part) by ELSA - European Lighthouse on Secure and Safe AI funded by the European Union under grant agreement No. 101070617. 
Views and opinions expressed are however those of the author(s) only and do not necessarily reflect those of any of these funding agencies, European Union or European Commission. 
We thank the Prolific participants and our volunteer pilot participants. We thank Alan Liu and Kartik Chandra for their invaluable input on designing our web-based server.

\bibliography{AIES/aies}

\begin{thebibliography}{113}
\providecommand{\natexlab}[1]{#1}

\bibitem[{EUA(2023)}]{EUAIAct}
 2023.
\newblock “Amendments adopted by the European Parliament on 14 June 2023 on
  the proposal for a regulation of the European Parliament and of the Council
  on laying down harmonised rules on artificial intelligence (Artificial
  Intelligence Act) and amending certain Union legislative acts”
  (COM(2021)0206 – C9-0146/2021 – 2021/0106(COD).

\bibitem[{Amodei et~al.(2016)Amodei, Olah, Steinhardt, Christiano, Schulman,
  and Man{\'e}}]{amodei2016concrete}
Amodei, D.; Olah, C.; Steinhardt, J.; Christiano, P.; Schulman, J.; and
  Man{\'e}, D. 2016.
\newblock Concrete problems in AI safety.
\newblock \emph{arXiv preprint arXiv:1606.06565}.

\bibitem[{Antoran et~al.(2020)Antoran, Bhatt, Adel, Weller, and
  Hern{\'a}ndez-Lobato}]{antoran2020getting}
Antoran, J.; Bhatt, U.; Adel, T.; Weller, A.; and Hern{\'a}ndez-Lobato, J.~M.
  2020.
\newblock Getting a CLUE: A Method for Explaining Uncertainty Estimates.
\newblock In \emph{International Conference on Learning Representations}.

\bibitem[{Augustin et~al.(2017)Augustin, Venanzi, Rogers, and
  Jennings}]{augustin2017bayesian}
Augustin, A.; Venanzi, M.; Rogers, A.; and Jennings, N.~R. 2017.
\newblock Bayesian Aggregation of Categorical Distributions with Applications
  in Crowdsourcing.
\newblock In \emph{IJCAI}, 1411--1417.

\bibitem[{Babbar, Bhatt, and Weller(2022)}]{babbar2022utility}
Babbar, V.; Bhatt, U.; and Weller, A. 2022.
\newblock On the Utility of Prediction Sets in Human-AI Teams.
\newblock In Raedt, L.~D., ed., \emph{Proceedings of the Thirty-First
  International Joint Conference on Artificial Intelligence, {IJCAI-22}},
  2457--2463. International Joint Conferences on Artificial Intelligence
  Organization.
\newblock Main Track.

\bibitem[{Bakker et~al.(2021)Bakker, Tu, Gummadi, Pentland, Varshney, and
  Weller}]{bakker2021beyond}
Bakker, M.~A.; Tu, D.~P.; Gummadi, K.~P.; Pentland, A.~S.; Varshney, K.~R.; and
  Weller, A. 2021.
\newblock Beyond reasonable doubt: Improving fairness in budget-constrained
  decision making using confidence thresholds.
\newblock In \emph{Proceedings of the 2021 AAAI/ACM Conference on AI, Ethics,
  and Society}, 346--356.

\bibitem[{Bansal et~al.(2021{\natexlab{a}})Bansal, Nushi, Kamar, Horvitz, and
  Weld}]{bansal2021most}
Bansal, G.; Nushi, B.; Kamar, E.; Horvitz, E.; and Weld, D.~S.
  2021{\natexlab{a}}.
\newblock Is the most accurate {AI} the best teammate? optimizing {AI} for
  teamwork.
\newblock In \emph{Proceedings of the AAAI Conference on Artificial
  Intelligence}, volume~35, 11405--11414.

\bibitem[{Bansal et~al.(2021{\natexlab{b}})Bansal, Wu, Zhou, Fok, Nushi, Kamar,
  Ribeiro, and Weld}]{bansal2021does}
Bansal, G.; Wu, T.; Zhou, J.; Fok, R.; Nushi, B.; Kamar, E.; Ribeiro, M.~T.;
  and Weld, D. 2021{\natexlab{b}}.
\newblock Does the whole exceed its parts? {T}he effect of {AI} explanations on
  complementary team performance.
\newblock In \emph{Proceedings of the 2021 CHI Conference on Human Factors in
  Computing Systems}, 1--16.

\bibitem[{Barabas(2022)}]{barabas2022refusal}
Barabas, C. 2022.
\newblock Refusal in Data Ethics: Re-Imagining the Code Beneath the Code of
  Computation in the Carceral State.
\newblock \emph{Engaging Science, Technology, and Society}, 8(2): 35--57.

\bibitem[{Bastani, Bastani, and Sinchaisri(2022)}]{bastani2022improving}
Bastani, H.; Bastani, O.; and Sinchaisri, P. 2022.
\newblock Improving Human Decision-Making with Machine Learning.
\newblock In \emph{Academy of Management Proceedings}, volume 2022, 17725.
  Academy of Management Briarcliff Manor, NY 10510.

\bibitem[{Bastani and Bayati(2020)}]{bastani2020online}
Bastani, H.; and Bayati, M. 2020.
\newblock Online decision making with high-dimensional covariates.
\newblock \emph{Operations Research}, 68(1): 276--294.

\bibitem[{Bates et~al.(2021)Bates, Angelopoulos, Lei, Malik, and
  Jordan}]{bates2021distribution}
Bates, S.; Angelopoulos, A.; Lei, L.; Malik, J.; and Jordan, M. 2021.
\newblock Distribution-free, risk-controlling prediction sets.
\newblock \emph{Journal of the ACM (JACM)}, 68(6): 1--34.

\bibitem[{Battleday, Peterson, and Griffiths(2020)}]{battleday2020capturing}
Battleday, R.~M.; Peterson, J.~C.; and Griffiths, T.~L. 2020.
\newblock Capturing human categorization of natural images by combining deep
  networks and cognitive models.
\newblock \emph{Nature communications}, 11(1): 1--14.

\bibitem[{Beyer et~al.(2020)Beyer, H{\'{e}}naff, Kolesnikov, Zhai, and van~den
  Oord}]{imagenetReaLH}
Beyer, L.; H{\'{e}}naff, O.~J.; Kolesnikov, A.; Zhai, X.; and van~den Oord, A.
  2020.
\newblock Are we done with ImageNet?
\newblock \emph{CoRR}, abs/2006.07159.

\bibitem[{Bhatt et~al.(2021)Bhatt, Antor\'{a}n, Zhang, Liao, Sattigeri,
  Fogliato, Melan\c{c}on, Krishnan, Stanley, Tickoo, Nachman, Chunara,
  Srikumar, Weller, and Xiang}]{bhatt2021uncertainty}
Bhatt, U.; Antor\'{a}n, J.; Zhang, Y.; Liao, Q.~V.; Sattigeri, P.; Fogliato,
  R.; Melan\c{c}on, G.; Krishnan, R.; Stanley, J.; Tickoo, O.; Nachman, L.;
  Chunara, R.; Srikumar, M.; Weller, A.; and Xiang, A. 2021.
\newblock Uncertainty as a {F}orm of {T}ransparency: {M}easuring,
  {C}ommunicating, and {U}sing {U}ncertainty.
\newblock In \emph{Proceedings of the 2021 AAAI/ACM Conference on AI, Ethics,
  and Society}, 401–413. New York, NY, USA.
\newblock ISBN 9781450384735.

\bibitem[{Biden(2023)}]{EO14110}
Biden, J.~R. 2023.
\newblock \emph{{Executive Order 14110, Executive Order on the Safe, Secure,
  and Trustworthy Development and Use of Artificial Intelligence}}.
\newblock The White House.

\bibitem[{Bommasani et~al.(2021)Bommasani, Hudson, Adeli, Altman, Arora, von
  Arx, Bernstein, Bohg, Bosselut, Brunskill, Brynjolfsson, Buch, Card,
  Castellon, Chatterji, Chen, Creel, Davis, Demszky, Donahue, Doumbouya,
  Durmus, Ermon, Etchemendy, Ethayarajh, Fei{-}Fei, Finn, Gale, Gillespie,
  Goel, Goodman, Grossman, Guha, Hashimoto, Henderson, Hewitt, Ho, Hong, Hsu,
  Huang, Icard, Jain, Jurafsky, Kalluri, Karamcheti, Keeling, Khani, Khattab,
  Koh, Krass, Krishna, Kuditipudi, and et~al.}]{foundationModels}
Bommasani, R.; Hudson, D.~A.; Adeli, E.; Altman, R.~B.; Arora, S.; von Arx, S.;
  Bernstein, M.~S.; Bohg, J.; Bosselut, A.; Brunskill, E.; Brynjolfsson, E.;
  Buch, S.; Card, D.; Castellon, R.; Chatterji, N.~S.; Chen, A.~S.; Creel, K.;
  Davis, J.~Q.; Demszky, D.; Donahue, C.; Doumbouya, M.; Durmus, E.; Ermon, S.;
  Etchemendy, J.; Ethayarajh, K.; Fei{-}Fei, L.; Finn, C.; Gale, T.; Gillespie,
  L.; Goel, K.; Goodman, N.~D.; Grossman, S.; Guha, N.; Hashimoto, T.;
  Henderson, P.; Hewitt, J.; Ho, D.~E.; Hong, J.; Hsu, K.; Huang, J.; Icard,
  T.; Jain, S.; Jurafsky, D.; Kalluri, P.; Karamcheti, S.; Keeling, G.; Khani,
  F.; Khattab, O.; Koh, P.~W.; Krass, M.~S.; Krishna, R.; Kuditipudi, R.; and
  et~al. 2021.
\newblock On the Opportunities and Risks of Foundation Models.
\newblock \emph{CoRR}, abs/2108.07258.

\bibitem[{Bondi et~al.(2022)Bondi, Koster, Sheahan, Chadwick, Bachrach, Cemgil,
  Paquet, and Dvijotham}]{bondi2021role}
Bondi, E.; Koster, R.; Sheahan, H.; Chadwick, M.; Bachrach, Y.; Cemgil, T.;
  Paquet, U.; and Dvijotham, K. 2022.
\newblock Role of {H}uman-{AI} {I}nteraction in {S}elective {P}rediction.
\newblock \emph{Proceedings of the AAAI Conference on Artificial Intelligence}.

\bibitem[{Bordt and Von~Luxburg(2022)}]{bordt2022bandit}
Bordt, S.; and Von~Luxburg, U. 2022.
\newblock A Bandit Model for Human-Machine Decision Making with Private
  Information and Opacity.
\newblock In \emph{International Conference on Artificial Intelligence and
  Statistics}, 7300--7319. PMLR.

\bibitem[{Briggs et~al.(2008)Briggs, Flynn, Worthington, Rennie, and
  McKinstry}]{briggs2008role}
Briggs, G.~M.; Flynn, P.~A.; Worthington, M.; Rennie, I.; and McKinstry, C.
  2008.
\newblock The role of specialist neuroradiology second opinion reporting: is
  there added value?
\newblock \emph{Clinical radiology}, 63(7): 791--795.

\bibitem[{Brundage et~al.(2018)Brundage, Avin, Clark, Toner, Eckersley,
  Garfinkel, Dafoe, Scharre, Zeitzoff, Filar et~al.}]{brundage2018malicious}
Brundage, M.; Avin, S.; Clark, J.; Toner, H.; Eckersley, P.; Garfinkel, B.;
  Dafoe, A.; Scharre, P.; Zeitzoff, T.; Filar, B.; et~al. 2018.
\newblock The malicious use of artificial intelligence: Forecasting,
  prevention, and mitigation.
\newblock \emph{arXiv preprint arXiv:1802.07228}.

\bibitem[{Bu{\c{c}}inca et~al.(2020)Bu{\c{c}}inca, Lin, Gajos, and
  Glassman}]{buccinca2020proxy}
Bu{\c{c}}inca, Z.; Lin, P.; Gajos, K.~Z.; and Glassman, E.~L. 2020.
\newblock Proxy tasks and subjective measures can be misleading in evaluating
  explainable AI systems.
\newblock In \emph{Proceedings of the 25th International Conference on
  Intelligent User Interfaces}, 454--464.

\bibitem[{Bu{\c{c}}inca, Malaya, and Gajos(2021)}]{buccinca2021trust}
Bu{\c{c}}inca, Z.; Malaya, M.~B.; and Gajos, K.~Z. 2021.
\newblock To trust or to think: cognitive forcing functions can reduce
  overreliance on AI in AI-assisted decision-making.
\newblock \emph{Proceedings of the ACM on Human-Computer Interaction},
  5(CSCW1): 1--21.

\bibitem[{Bu{\c{c}}inca et~al.(2024)Bu{\c{c}}inca, Swaroop, Paluch, Murphy, and
  Gajos}]{buccinca2024towards}
Bu{\c{c}}inca, Z.; Swaroop, S.; Paluch, A.~E.; Murphy, S.~A.; and Gajos, K.~Z.
  2024.
\newblock Towards Optimizing Human-Centric Objectives in AI-Assisted
  Decision-Making With Offline Reinforcement Learning.
\newblock \emph{arXiv preprint arXiv:2403.05911}.

\bibitem[{Bussone, Stumpf, and O'Sullivan(2015)}]{bussone2015role}
Bussone, A.; Stumpf, S.; and O'Sullivan, D. 2015.
\newblock The role of explanations on trust and reliance in clinical decision
  support systems.
\newblock In \emph{2015 international conference on healthcare informatics},
  160--169. IEEE.

\bibitem[{Camburu et~al.(2018)Camburu, Rockt{\"a}schel, Lukasiewicz, and
  Blunsom}]{camburu2018snli}
Camburu, O.-M.; Rockt{\"a}schel, T.; Lukasiewicz, T.; and Blunsom, P. 2018.
\newblock e-snli: Natural language inference with natural language
  explanations.
\newblock \emph{Advances in Neural Information Processing Systems}, 31.

\bibitem[{Charusaie et~al.(2022)Charusaie, Mozannar, Sontag, and
  Samadi}]{charusaie2022sample}
Charusaie, M.-A.; Mozannar, H.; Sontag, D.; and Samadi, S. 2022.
\newblock Sample Efficient Learning of Predictors that Complement Humans.
\newblock In \emph{International Conference on Machine Learning}, 2972--3005.
  PMLR.

\bibitem[{Chen et~al.(2022)Chen, Li, Kim, Plumb, and
  Talwalkar}]{chen2022interpretable}
Chen, V.; Li, J.; Kim, J.~S.; Plumb, G.; and Talwalkar, A. 2022.
\newblock Interpretable machine learning: Moving from mythos to diagnostics.
\newblock \emph{Queue}, 19(6): 28--56.

\bibitem[{Chen et~al.(2023)Chen, Liao, Wortman~Vaughan, and
  Bansal}]{chen2023understanding}
Chen, V.; Liao, Q.~V.; Wortman~Vaughan, J.; and Bansal, G. 2023.
\newblock Understanding the role of human intuition on reliance in human-AI
  decision-making with explanations.
\newblock \emph{Proceedings of the ACM on Human-Computer Interaction},
  7(CSCW2): 1--32.

\bibitem[{Chow(1957)}]{chow1957optimum}
Chow, C.-K. 1957.
\newblock An optimum character recognition system using decision functions.
\newblock \emph{IRE Transactions on Electronic Computers}, (4): 247--254.

\bibitem[{Collins et~al.(2023)Collins, Barker, Zarlenga, Raman, Bhatt, Jamnik,
  Sucholutsky, Weller, and Dvijotham}]{collins2023human}
Collins, K.~M.; Barker, M.; Zarlenga, M.~E.; Raman, N.; Bhatt, U.; Jamnik, M.;
  Sucholutsky, I.; Weller, A.; and Dvijotham, K. 2023.
\newblock Human Uncertainty in Concept-Based AI Systems.
\newblock arXiv:2303.12872.

\bibitem[{Collins, Bhatt, and Weller(2022)}]{softLabelElicitingLearning2022}
Collins, K.~M.; Bhatt, U.; and Weller, A. 2022.
\newblock Eliciting and Learning with Soft Labels from Every Annotator.
\newblock In \emph{Proceedings of the AAAI Conference on Human Computation and
  Crowdsourcing (HCOMP)}, volume~10.

\bibitem[{Collins et~al.(2024{\natexlab{a}})Collins, Jiang, Frieder, Wong,
  Zilka, Bhatt, Lukasiewicz, Wu, Tenenbaum, Hart
  et~al.}]{collins2023evaluating}
Collins, K.~M.; Jiang, A.~Q.; Frieder, S.; Wong, L.; Zilka, M.; Bhatt, U.;
  Lukasiewicz, T.; Wu, Y.; Tenenbaum, J.~B.; Hart, W.; et~al.
  2024{\natexlab{a}}.
\newblock Evaluating language models for mathematics through interactions.
\newblock \emph{Proceedings of the National Academy of Sciences}, 121(24):
  e2318124121.

\bibitem[{Collins et~al.(2024{\natexlab{b}})Collins, Sucholutsky, Bhatt,
  Chandra, Wong, Lee, Zhang, Zhi-Xuan, Ho, Mansinghka
  et~al.}]{collins2024building}
Collins, K.~M.; Sucholutsky, I.; Bhatt, U.; Chandra, K.; Wong, L.; Lee, M.;
  Zhang, C.~E.; Zhi-Xuan, T.; Ho, M.; Mansinghka, V.; et~al.
  2024{\natexlab{b}}.
\newblock Building machines that learn and think with people.
\newblock \emph{Nature Human Behaviour}, 8(10): 1851--1863.

\bibitem[{Cortes et~al.(2018)Cortes, DeSalvo, Gentile, Mohri, and
  Yang}]{cortes2018online}
Cortes, C.; DeSalvo, G.; Gentile, C.; Mohri, M.; and Yang, S. 2018.
\newblock Online learning with abstention.
\newblock In \emph{International conference on machine learning}, 1059--1067.
  PMLR.

\bibitem[{Cortes, DeSalvo, and Mohri(2016)}]{cortes2016learning}
Cortes, C.; DeSalvo, G.; and Mohri, M. 2016.
\newblock Learning with rejection.
\newblock In \emph{International Conference on Algorithmic Learning Theory},
  67--82. Springer.

\bibitem[{Dawid and Skene(1979)}]{dawid1979maximum}
Dawid, A.~P.; and Skene, A.~M. 1979.
\newblock Maximum likelihood estimation of observer error-rates using the EM
  algorithm.
\newblock \emph{Journal of the Royal Statistical Society: Series C (Applied
  Statistics)}, 28(1): 20--28.

\bibitem[{De-Arteaga, Dubrawski, and Chouldechova(2018)}]{de2018learning}
De-Arteaga, M.; Dubrawski, A.; and Chouldechova, A. 2018.
\newblock Learning under selective labels in the presence of expert
  consistency.
\newblock \emph{arXiv preprint arXiv:1807.00905}.

\bibitem[{Ehsan et~al.(2018)Ehsan, Harrison, Chan, and
  Riedl}]{ehsan2018rationalization}
Ehsan, U.; Harrison, B.; Chan, L.; and Riedl, M.~O. 2018.
\newblock Rationalization: A neural machine translation approach to generating
  natural language explanations.
\newblock In \emph{Proceedings of the 2018 AAAI/ACM Conference on AI, Ethics,
  and Society}, 81--87.

\bibitem[{Erramilli(1996)}]{erramilli1996nationality}
Erramilli, M.~K. 1996.
\newblock Nationality and subsidiary ownership patterns in multinational
  corporations.
\newblock \emph{Journal of International Business Studies}, 27: 225--248.

\bibitem[{Gabriel et~al.(2024)Gabriel, Manzini, Keeling, Hendricks, Rieser,
  Iqbal, Toma{\v{s}}ev, Ktena, Kenton, Rodriguez et~al.}]{gabriel2024ethics}
Gabriel, I.; Manzini, A.; Keeling, G.; Hendricks, L.~A.; Rieser, V.; Iqbal, H.;
  Toma{\v{s}}ev, N.; Ktena, I.; Kenton, Z.; Rodriguez, M.; et~al. 2024.
\newblock The Ethics of Advanced AI Assistants.
\newblock \emph{arXiv preprint arXiv:2404.16244}.

\bibitem[{Gao et~al.(2021)Gao, Saar-Tsechansky, De-Arteaga, Han, Lee, and
  Lease}]{gao2021human}
Gao, R.; Saar-Tsechansky, M.; De-Arteaga, M.; Han, L.; Lee, M.~K.; and Lease,
  M. 2021.
\newblock Human-{AI} Collaboration with Bandit Feedback.
\newblock In \emph{Proceedings of the Thirtieth International Joint Conference
  on Artificial Intelligence, {IJCAI-21}}, 1722--1728. International Joint
  Conferences on Artificial Intelligence Organization.

\bibitem[{Gao et~al.(2023)Gao, Saar-Tsechansky, De-Arteaga, Han, Sun, Lee, and
  Lease}]{gao2023learning}
Gao, R.; Saar-Tsechansky, M.; De-Arteaga, M.; Han, L.; Sun, W.; Lee, M.~K.; and
  Lease, M. 2023.
\newblock Learning Complementary Policies for Human-AI Teams.
\newblock \emph{arXiv preprint arXiv:2302.02944}.

\bibitem[{Geifman and El-Yaniv(2017)}]{geifman2017selective}
Geifman, Y.; and El-Yaniv, R. 2017.
\newblock Selective classification for deep neural networks.
\newblock \emph{Advances in neural information processing systems}, 30.

\bibitem[{Gordon and Mugar(2020)}]{gordon_meaningful_2020}
Gordon, E.; and Mugar, G. 2020.
\newblock \emph{Meaningful inefficiencies: designing for public value in an age
  of digital expediency}.
\newblock New York, NY: Oxford University Press.
\newblock ISBN 978-0-19-087017-1 978-0-19-087016-4 978-0-19-087015-7.

\bibitem[{Gordon et~al.(2022)Gordon, Lam, Park, Patel, Hancock, Hashimoto, and
  Bernstein}]{gordon2022jury}
Gordon, M.~L.; Lam, M.~S.; Park, J.~S.; Patel, K.; Hancock, J.; Hashimoto, T.;
  and Bernstein, M.~S. 2022.
\newblock Jury learning: Integrating dissenting voices into machine learning
  models.
\newblock In \emph{CHI Conference on Human Factors in Computing Systems},
  1--19.

\bibitem[{Gordon et~al.(2021)Gordon, Zhou, Patel, Hashimoto, and
  Bernstein}]{gordon2021disagreement}
Gordon, M.~L.; Zhou, K.; Patel, K.; Hashimoto, T.; and Bernstein, M.~S. 2021.
\newblock The disagreement deconvolution: Bringing machine learning performance
  metrics in line with reality.
\newblock In \emph{Proceedings of the 2021 CHI Conference on Human Factors in
  Computing Systems}, 1--14.

\bibitem[{Green and Chen(2019)}]{green2019principles}
Green, B.; and Chen, Y. 2019.
\newblock The principles and limits of algorithm-in-the-loop decision making.
\newblock \emph{Proceedings of the ACM on Human-Computer Interaction}, 3(CSCW):
  1--24.

\bibitem[{Guan and Jiang(2018)}]{guan2018nonparametric}
Guan, M.; and Jiang, H. 2018.
\newblock Nonparametric stochastic contextual bandits.
\newblock In \emph{Proceedings of the AAAI Conference on Artificial
  Intelligence}, volume~32.

\bibitem[{Harrell et~al.(1982)Harrell, Califf, Pryor, Lee, and
  Rosati}]{harrell1982evaluating}
Harrell, F.~E.; Califf, R.~M.; Pryor, D.~B.; Lee, K.~L.; and Rosati, R.~A.
  1982.
\newblock Evaluating the yield of medical tests.
\newblock \emph{Jama}, 247(18): 2543--2546.

\bibitem[{He et~al.(2016)He, Zhang, Ren, and Sun}]{he2016deep}
He, K.; Zhang, X.; Ren, S.; and Sun, J. 2016.
\newblock Deep residual learning for image recognition.
\newblock In \emph{CVPR}, 770--778.

\bibitem[{Hemmer et~al.(2022)Hemmer, Schellhammer, V{\"o}ssing, Jakubik, and
  Satzger}]{hemmer2022forming}
Hemmer, P.; Schellhammer, S.; V{\"o}ssing, M.; Jakubik, J.; and Satzger, G.
  2022.
\newblock Forming Effective Human-{AI} Teams: Building Machine Learning Models
  that Complement the Capabilities of Multiple Experts.
\newblock In \emph{Proceedings of the Thirtieth International Conference on
  International Joint Conferences on Artificial Intelligence}.

\bibitem[{Hemmer et~al.(2023)Hemmer, Westphal, Schemmer, Vetter, V{\"o}ssing,
  and Satzger}]{hemmer2023human}
Hemmer, P.; Westphal, M.; Schemmer, M.; Vetter, S.; V{\"o}ssing, M.; and
  Satzger, G. 2023.
\newblock Human-AI Collaboration: The Effect of AI Delegation on Human Task
  Performance and Task Satisfaction.
\newblock In \emph{Proceedings of the 28th International Conference on
  Intelligent User Interfaces}, 453--463.

\bibitem[{Hendrycks et~al.(2020)Hendrycks, Burns, Basart, Zou, Mazeika, Song,
  and Steinhardt}]{hendrycks2020measuring}
Hendrycks, D.; Burns, C.; Basart, S.; Zou, A.; Mazeika, M.; Song, D.; and
  Steinhardt, J. 2020.
\newblock Measuring Massive Multitask Language Understanding.
\newblock In \emph{International Conference on Learning Representations}.

\bibitem[{Hullman et~al.(2018)Hullman, Qiao, Correll, Kale, and
  Kay}]{hullman2018pursuit}
Hullman, J.; Qiao, X.; Correll, M.; Kale, A.; and Kay, M. 2018.
\newblock In pursuit of error: A survey of uncertainty visualization
  evaluation.
\newblock \emph{IEEE transactions on visualization and computer graphics},
  25(1): 903--913.

\bibitem[{James et~al.(2023)James, Nagpal, Heller, and
  Ustun}]{james2023participatory}
James, H.; Nagpal, C.; Heller, K.~A.; and Ustun, B. 2023.
\newblock Participatory Personalization in Classification.
\newblock In \emph{Thirty-seventh Conference on Neural Information Processing
  Systems}.

\bibitem[{Jeyakumar et~al.(2020)Jeyakumar, Noor, Cheng, Garcia, and
  Srivastava}]{jeyakumar2020can}
Jeyakumar, J.~V.; Noor, J.; Cheng, Y.-H.; Garcia, L.; and Srivastava, M. 2020.
\newblock How Can I Explain This to You? An Empirical Study of Deep Neural
  Network Explanation Methods.
\newblock \emph{Advances in Neural Information Processing Systems}, 33.

\bibitem[{Kahn~Jr(1994)}]{kahn1994artificial}
Kahn~Jr, C.~E. 1994.
\newblock Artificial intelligence in radiology: decision support systems.
\newblock \emph{Radiographics}, 14(4): 849--861.

\bibitem[{Kalis, Kaiser, and Mojzisch(2013)}]{kalis2013we}
Kalis, A.; Kaiser, S.; and Mojzisch, A. 2013.
\newblock Why we should talk about option generation in decision-making
  research.
\newblock \emph{Frontiers in psychology}, 4: 555.

\bibitem[{Keen(1980)}]{keen1980decision}
Keen, P.~G. 1980.
\newblock Decision support systems: a research perspective.
\newblock In \emph{Decision support systems: Issues and challenges: Proceedings
  of an international task force meeting}, 23--44.

\bibitem[{Keswani, Lease, and Kenthapadi(2021)}]{keswani2021towards}
Keswani, V.; Lease, M.; and Kenthapadi, K. 2021.
\newblock Towards unbiased and accurate deferral to multiple experts.
\newblock In \emph{Proceedings of the 2021 AAAI/ACM Conference on AI, Ethics,
  and Society}, 154--165.

\bibitem[{Kim, Rudin, and Shah(2014)}]{kim2014bayesian}
Kim, B.; Rudin, C.; and Shah, J.~A. 2014.
\newblock The bayesian case model: A generative approach for case-based
  reasoning and prototype classification.
\newblock In \emph{Advances in neural information processing systems},
  1952--1960.

\bibitem[{Kirk et~al.(2024)Kirk, Whitefield, R{\"o}ttger, Bean, Margatina,
  Ciro, Mosquera, Bartolo, Williams, He et~al.}]{kirk2024prism}
Kirk, H.~R.; Whitefield, A.; R{\"o}ttger, P.; Bean, A.; Margatina, K.; Ciro,
  J.; Mosquera, R.; Bartolo, M.; Williams, A.; He, H.; et~al. 2024.
\newblock The PRISM Alignment Project: What Participatory, Representative and
  Individualised Human Feedback Reveals About the Subjective and Multicultural
  Alignment of Large Language Models.
\newblock \emph{arXiv preprint arXiv:2404.16019}.

\bibitem[{Krizhevsky(2009)}]{krizhevsky2009learning}
Krizhevsky, A. 2009.
\newblock Learning Multiple Layers of Features from Tiny Images.
\newblock \emph{Master's thesis, University of Toronto}.

\bibitem[{Lai et~al.(2022)Lai, Carton, Bhatnagar, Liao, Zhang, and
  Tan}]{lai2022human}
Lai, V.; Carton, S.; Bhatnagar, R.; Liao, Q.~V.; Zhang, Y.; and Tan, C. 2022.
\newblock Human-{AI} Collaboration via Conditional Delegation: A Case Study of
  Content Moderation.
\newblock In \emph{CHI Conference on Human Factors in Computing Systems},
  1--18.

\bibitem[{Lai et~al.(2023)Lai, Chen, Smith-Renner, Liao, and
  Tan}]{lai2021towards}
Lai, V.; Chen, C.; Smith-Renner, A.; Liao, Q.~V.; and Tan, C. 2023.
\newblock Towards a Science of Human-AI Decision Making: An Overview of Design
  Space in Empirical Human-Subject Studies.
\newblock In \emph{Proceedings of the 2023 ACM Conference on Fairness,
  Accountability, and Transparency}, 1369--1385.

\bibitem[{Laidlaw and Russell(2021)}]{laidlaw2021uncertain}
Laidlaw, C.; and Russell, S. 2021.
\newblock Uncertain Decisions Facilitate Better Preference Learning.
\newblock \emph{Advances in Neural Information Processing Systems}, 34:
  15070--15083.

\bibitem[{Lee et~al.(2023)Lee, Srivastava, Hardy, Thickstun, Durmus, Paranjape,
  Gerard-Ursin, Li, Ladhak, Rong, Wang, Kwon, Park, Cao, Lee, Bommasani,
  Bernstein, and Liang}]{lee2023evaluating}
Lee, M.; Srivastava, M.; Hardy, A.; Thickstun, J.; Durmus, E.; Paranjape, A.;
  Gerard-Ursin, I.; Li, X.~L.; Ladhak, F.; Rong, F.; Wang, R.~E.; Kwon, M.;
  Park, J.~S.; Cao, H.; Lee, T.; Bommasani, R.; Bernstein, M.~S.; and Liang, P.
  2023.
\newblock Evaluating Human-Language Model Interaction.
\newblock \emph{Transactions on Machine Learning Research}.

\bibitem[{Li et~al.(2010)Li, Chu, Langford, and Schapire}]{li2010contextual}
Li, L.; Chu, W.; Langford, J.; and Schapire, R.~E. 2010.
\newblock A contextual-bandit approach to personalized news article
  recommendation.
\newblock In \emph{Proceedings of the 19th international conference on World
  wide web}, 661--670.

\bibitem[{Lichtenstein, Fischhoff, and
  Phillips(1977)}]{lichtenstein1977calibration}
Lichtenstein, S.; Fischhoff, B.; and Phillips, L.~D. 1977.
\newblock Calibration of probabilities: The state of the art.
\newblock \emph{Decision making and change in human affairs}, 275--324.

\bibitem[{Ma et~al.(2023)Ma, Lei, Wang, Zheng, Shi, Yin, and Ma}]{ma2023should}
Ma, S.; Lei, Y.; Wang, X.; Zheng, C.; Shi, C.; Yin, M.; and Ma, X. 2023.
\newblock Who should i trust: Ai or myself? leveraging human and ai correctness
  likelihood to promote appropriate trust in ai-assisted decision-making.
\newblock In \emph{Proceedings of the 2023 CHI Conference on Human Factors in
  Computing Systems}, 1--19.

\bibitem[{Madras, Pitassi, and Zemel(2018)}]{madras2018predict}
Madras, D.; Pitassi, T.; and Zemel, R. 2018.
\newblock Predict responsibly: improving fairness and accuracy by learning to
  defer.
\newblock \emph{Advances in Neural Information Processing Systems}, 31.

\bibitem[{Mozannar et~al.(2022)Mozannar, Bansal, Fourney, and
  Horvitz}]{mozannar2022reading}
Mozannar, H.; Bansal, G.; Fourney, A.; and Horvitz, E. 2022.
\newblock Reading Between the Lines: Modeling User Behavior and Costs in
  AI-Assisted Programming.
\newblock \emph{arXiv preprint arXiv:2210.14306}.

\bibitem[{Mozannar et~al.(2023)Mozannar, Lee, Wei, Sattigeri, Das, and
  Sontag}]{NEURIPS2023_61355b9c}
Mozannar, H.; Lee, J.; Wei, D.; Sattigeri, P.; Das, S.; and Sontag, D. 2023.
\newblock Effective Human-AI Teams via Learned Natural Language Rules and
  Onboarding.
\newblock In Oh, A.; Naumann, T.; Globerson, A.; Saenko, K.; Hardt, M.; and
  Levine, S., eds., \emph{Advances in Neural Information Processing Systems},
  volume~36, 30466--30498. Curran Associates, Inc.

\bibitem[{Mozannar, Satyanarayan, and Sontag(2022)}]{mozannar2022teaching}
Mozannar, H.; Satyanarayan, A.; and Sontag, D. 2022.
\newblock Teaching Humans When To Defer to a Classifier via Exemplars.
\newblock In \emph{Proceedings of the AAAI Conference on Artificial
  Intelligence}, volume~36, 5323--5331.

\bibitem[{Mozannar and Sontag(2020)}]{mozannar2020consistent}
Mozannar, H.; and Sontag, D. 2020.
\newblock Consistent estimators for learning to defer to an expert.
\newblock In \emph{International Conference on Machine Learning}, 7076--7087.
  PMLR.

\bibitem[{Mylonakis et~al.(2000)Mylonakis, Paliou, Greenbough, Flaningan,
  Letvin, and Rich}]{mylonakis2000report}
Mylonakis, E.; Paliou, M.; Greenbough, T.~C.; Flaningan, T.~P.; Letvin, N.~L.;
  and Rich, J.~D. 2000.
\newblock Report of a false-positive HIV test result and the potential use of
  additional tests in establishing HIV serostatus.
\newblock \emph{Archives of internal medicine}, 160(15): 2386--2388.

\bibitem[{Nakano et~al.(2021)Nakano, Hilton, Balaji, Wu, Ouyang, Kim, Hesse,
  Jain, Kosaraju, Saunders, Jiang, Cobbe, Eloundou, Krueger, Button, Knight,
  Chess, and Schulman}]{webgpt}
Nakano, R.; Hilton, J.; Balaji, S.; Wu, J.; Ouyang, L.; Kim, C.; Hesse, C.;
  Jain, S.; Kosaraju, V.; Saunders, W.; Jiang, X.; Cobbe, K.; Eloundou, T.;
  Krueger, G.; Button, K.; Knight, M.; Chess, B.; and Schulman, J. 2021.
\newblock WebGPT: Browser-assisted question-answering with human feedback.
\newblock \emph{CoRR}, abs/2112.09332.

\bibitem[{Noti and Chen(2023)}]{noti2022learning}
Noti, G.; and Chen, Y. 2023.
\newblock Learning when to advise human decision makers.
\newblock In \emph{Proceedings of the Thirty-Second International Joint
  Conference on Artificial Intelligence}, 3038--3048.

\bibitem[{O'Hagan et~al.(2006)O'Hagan, Buck, Daneshkhah, Eiser, Garthwaite,
  Jenkinson, Oakley, and Rakow}]{uncertainJudgments}
O'Hagan, A.; Buck, C.~E.; Daneshkhah, A.; Eiser, J.~R.; Garthwaite, P.~H.;
  Jenkinson, D.~J.; Oakley, J.~E.; and Rakow, T. 2006.
\newblock \emph{Uncertain Judgements: Eliciting Expert Probabilities}.
\newblock Chichester: John Wiley.

\bibitem[{Okati, De, and Rodriguez(2021)}]{okati2021differentiable}
Okati, N.; De, A.; and Rodriguez, M. 2021.
\newblock Differentiable learning under triage.
\newblock \emph{Advances in Neural Information Processing Systems}, 34:
  9140--9151.

\bibitem[{Ouyang et~al.(2022)Ouyang, Wu, Jiang, Almeida, Wainwright, Mishkin,
  Zhang, Agarwal, Slama, Ray, Schulman, Hilton, Kelton, Miller, Simens, Askell,
  Welinder, Christiano, Leike, and Lowe}]{instructgpt}
Ouyang, L.; Wu, J.; Jiang, X.; Almeida, D.; Wainwright, C.~L.; Mishkin, P.;
  Zhang, C.; Agarwal, S.; Slama, K.; Ray, A.; Schulman, J.; Hilton, J.; Kelton,
  F.; Miller, L.; Simens, M.; Askell, A.; Welinder, P.; Christiano, P.; Leike,
  J.; and Lowe, R. 2022.
\newblock Training language models to follow instructions with human feedback.

\bibitem[{Palan and Schitter(2018)}]{palan2018prolific}
Palan, S.; and Schitter, C. 2018.
\newblock Prolific. ac—A subject pool for online experiments.
\newblock \emph{Journal of Behavioral and Experimental Finance}, 17: 22--27.

\bibitem[{Peterson et~al.(2019)Peterson, Battleday, Griffiths, and
  Russakovsky}]{peterson2019human}
Peterson, J.~C.; Battleday, R.~M.; Griffiths, T.~L.; and Russakovsky, O. 2019.
\newblock Human uncertainty makes classification more robust.
\newblock In \emph{Proceedings of the IEEE/CVF International Conference on
  Computer Vision}, 9617--9626.

\bibitem[{Phillips-Wren(2012)}]{phillips2012ai}
Phillips-Wren, G. 2012.
\newblock AI tools in decision making support systems: a review.
\newblock \emph{International Journal on Artificial Intelligence Tools},
  21(02): 1240005.

\bibitem[{Ribeiro, Singh, and Guestrin(2016)}]{ribeiro2016should}
Ribeiro, M.~T.; Singh, S.; and Guestrin, C. 2016.
\newblock " Why should i trust you?" Explaining the predictions of any
  classifier.
\newblock In \emph{Proceedings of the 22nd ACM SIGKDD international conference
  on knowledge discovery and data mining}, 1135--1144.

\bibitem[{Roda(2011)}]{roda2011human}
Roda, C.~E. 2011.
\newblock \emph{Human attention and its implications for human-computer
  interaction.}
\newblock Cambridge University Press.

\bibitem[{Scheife et~al.(2015)Scheife, Hines, Boyce, Chung, Momper, Sommer,
  Abernethy, Horn, Sklar, Wong et~al.}]{scheife2015consensus}
Scheife, R.~T.; Hines, L.~E.; Boyce, R.~D.; Chung, S.~P.; Momper, J.~D.;
  Sommer, C.~D.; Abernethy, D.~R.; Horn, J.~R.; Sklar, S.~J.; Wong, S.~K.;
  et~al. 2015.
\newblock Consensus recommendations for systematic evaluation of drug--drug
  interaction evidence for clinical decision support.
\newblock \emph{Drug safety}, 38(2): 197--206.

\bibitem[{Schvaneveldt et~al.(1985)Schvaneveldt, Durso, Goldsmith, Breen,
  Cooke, Tucker, and De~Maio}]{schvaneveldt1985measuring}
Schvaneveldt, R.~W.; Durso, F.~T.; Goldsmith, T.~E.; Breen, T.~J.; Cooke,
  N.~M.; Tucker, R.~G.; and De~Maio, J.~C. 1985.
\newblock Measuring the structure of expertise.
\newblock \emph{International journal of man-machine studies}, 23(6): 699--728.

\bibitem[{Seger and Peterson(2013)}]{seger2013categorization}
Seger, C.~A.; and Peterson, E.~J. 2013.
\newblock Categorization= decision making+ generalization.
\newblock \emph{Neuroscience \& Biobehavioral Reviews}, 37(7): 1187--1200.

\bibitem[{Spiegelhalter(2017)}]{spiegelhalter_risk_2017}
Spiegelhalter, D. 2017.
\newblock Risk and {{Uncertainty Communication}}.
\newblock 4(1): 31--60.

\bibitem[{Steyvers and Kumar(2022)}]{steyvers2022three}
Steyvers, M.; and Kumar, A. 2022.
\newblock Three Challenges for AI-Assisted Decision-Making.

\bibitem[{Straitouri et~al.(2022)Straitouri, Wang, Okati, and
  Rodriguez}]{straitouri2022provably}
Straitouri, E.; Wang, L.; Okati, N.; and Rodriguez, M.~G. 2022.
\newblock Provably improving expert predictions with conformal prediction.
\newblock \emph{arXiv preprint arXiv:2201.12006}.

\bibitem[{Sutton and Barto(2018)}]{sutton2018reinforcement}
Sutton, R.~S.; and Barto, A.~G. 2018.
\newblock \emph{Reinforcement learning: An introduction}.

\bibitem[{Swaroop et~al.(2024)Swaroop, Bu{\c{c}}inca, Gajos, and
  Doshi-Velez}]{swaroop2024accuracy}
Swaroop, S.; Bu{\c{c}}inca, Z.; Gajos, K.~Z.; and Doshi-Velez, F. 2024.
\newblock Accuracy-Time Tradeoffs in AI-Assisted Decision Making under Time
  Pressure.
\newblock In \emph{Proceedings of the 29th International Conference on
  Intelligent User Interfaces}, 138--154.

\bibitem[{Tejeda et~al.(2022)Tejeda, Kumar, Smyth, and Steyvers}]{tejeda2022ai}
Tejeda, H.; Kumar, A.; Smyth, P.; and Steyvers, M. 2022.
\newblock AI-assisted decision-making: A cognitive modeling approach to infer
  latent reliance strategies.
\newblock \emph{Computational Brain \& Behavior}, 1--18.

\bibitem[{Tekin and Tur{\u{g}}ay(2018)}]{tekin2018multi}
Tekin, C.; and Tur{\u{g}}ay, E. 2018.
\newblock Multi-objective contextual multi-armed bandit with a dominant
  objective.
\newblock \emph{IEEE Transactions on Signal Processing}, 66(14): 3799--3813.

\bibitem[{Turgay, Oner, and Tekin(2018)}]{turgay2018multi}
Turgay, E.; Oner, D.; and Tekin, C. 2018.
\newblock Multi-objective contextual bandit problem with similarity
  information.
\newblock In \emph{International Conference on Artificial Intelligence and
  Statistics}, 1673--1681. PMLR.

\bibitem[{Tversky and Kahneman(1996)}]{kahneman1996reality}
Tversky, A.; and Kahneman, D. 1996.
\newblock On the reality of cognitive illusions.
\newblock \emph{Psychological Review}, 103(3): 582--591.

\bibitem[{Uma, Almanea, and Poesio(2022)}]{uma2022scaling}
Uma, A.; Almanea, D.; and Poesio, M. 2022.
\newblock Scaling and Disagreements: Bias, Noise, and Ambiguity.
\newblock \emph{Frontiers in Artificial Intelligence}, 5.

\bibitem[{Uma et~al.(2020)Uma, Fornaciari, Hovy, Paun, Plank, and
  Poesio}]{Uma_Fornaciari_Hovy_Paun_Plank_Poesio_2020}
Uma, A.; Fornaciari, T.; Hovy, D.; Paun, S.; Plank, B.; and Poesio, M. 2020.
\newblock A Case for Soft Loss Functions.
\newblock \emph{Proceedings of the AAAI Conference on Human Computation and
  Crowdsourcing}, 8(1): 173--177.

\bibitem[{Ustun, Spangher, and Liu(2019)}]{ustun2019actionable}
Ustun, B.; Spangher, A.; and Liu, Y. 2019.
\newblock Actionable recourse in linear classification.
\newblock In \emph{Proceedings of the Conference on Fairness, Accountability,
  and Transparency}, 10--19.

\bibitem[{Vovk(1998)}]{vovk1998game}
Vovk, V. 1998.
\newblock A game of prediction with expert advice.
\newblock \emph{Journal of Computer and System Sciences}, 56(2): 153--173.

\bibitem[{Vovk, Gammerman, and Shafer(2005)}]{vovk2005}
Vovk, V.; Gammerman, A.; and Shafer, G. 2005.
\newblock \emph{Algorithmic Learning in a Random World}.
\newblock Springer.

\bibitem[{Wei et~al.(2022)Wei, Zhu, Cheng, Liu, Niu, and Liu}]{wei2022learning}
Wei, J.; Zhu, Z.; Cheng, H.; Liu, T.; Niu, G.; and Liu, Y. 2022.
\newblock Learning with Noisy Labels Revisited: A Study Using Real-World Human
  Annotations.
\newblock In \emph{International Conference on Learning Representations}.

\bibitem[{Whitehill et~al.(2009)Whitehill, Wu, Bergsma, Movellan, and
  Ruvolo}]{whitehill2009whoseVote}
Whitehill, J.; Wu, T.-f.; Bergsma, J.; Movellan, J.; and Ruvolo, P. 2009.
\newblock Whose vote should count more: Optimal integration of labels from
  labelers of unknown expertise.
\newblock \emph{Advances in neural information processing systems}, 22.

\bibitem[{Wilder, Horvitz, and Kamar(2021)}]{wilder2021learning}
Wilder, B.; Horvitz, E.; and Kamar, E. 2021.
\newblock Learning to complement humans.
\newblock In \emph{Proceedings of the Twenty-Ninth International Conference on
  International Joint Conferences on Artificial Intelligence}, 1526--1533.

\bibitem[{Wolczynski, Saar-Tsechansky, and Wang(2022)}]{wolczynski2022learning}
Wolczynski, N.; Saar-Tsechansky, M.; and Wang, T. 2022.
\newblock Learning to Advise Humans By Leveraging Algorithm Discretion.
\newblock \emph{arXiv preprint arXiv:2210.12849}.

\bibitem[{Yang et~al.(2023)Yang, Nachum, Du, Wei, Abbeel, and
  Schuurmans}]{yang2023foundation}
Yang, S.; Nachum, O.; Du, Y.; Wei, J.; Abbeel, P.; and Schuurmans, D. 2023.
\newblock Foundation Models for Decision Making: Problems, Methods, and
  Opportunities.
\newblock \emph{arXiv preprint arXiv:2303.04129}.

\bibitem[{Yu et~al.(2024)Yu, Moehring, Banerjee, Salz, Agarwal, and
  Rajpurkar}]{yu2024heterogeneity}
Yu, F.; Moehring, A.; Banerjee, O.; Salz, T.; Agarwal, N.; and Rajpurkar, P.
  2024.
\newblock Heterogeneity and predictors of the effects of AI assistance on
  radiologists.
\newblock \emph{Nature Medicine}, 30(3): 837--849.

\bibitem[{Zerilli, Bhatt, and Weller(2022)}]{zerilli2022transparency}
Zerilli, J.; Bhatt, U.; and Weller, A. 2022.
\newblock How transparency modulates trust in artificial intelligence.
\newblock \emph{Patterns}, 100455.

\bibitem[{Zhang, Liao, and Bellamy(2020)}]{zhang2020effect}
Zhang, Y.; Liao, Q.~V.; and Bellamy, R.~K. 2020.
\newblock Effect of confidence and explanation on accuracy and trust
  calibration in AI-assisted decision making.
\newblock In \emph{Proceedings of the 2020 conference on fairness,
  accountability, and transparency}, 295--305.

\bibitem[{Zuboff(2023)}]{zuboff2023age}
Zuboff, S. 2023.
\newblock The age of surveillance capitalism.
\newblock In \emph{Social theory re-wired}, 203--213. Routledge.

\end{thebibliography}



\clearpage

\appendix

We provide details on decision support policies, additional computational experiments with \texttt{Modiste}, and extensive information on our human subject experiments with \texttt{Modiste}.

\section{Comparison Against Prior Work}\label{appdx:comparison}

Most papers on human-AI collaboration have considered clever ways of abstaining from prediction on specific inputs~\citep{cortes2016learning,cortes2018online}, learning deferral functions based on multiple experts~\citep{vovk1998game,keswani2021towards}, or teaching decision-makers when to rely~\citep{mozannar2022teaching}.
There are also a number of papers from the HCI literature (see survey by~\citep{lai2021towards}) that evaluate the two-action setting of our formulation using a \emph{static} policy (e.g., always showing the ML model prediction or always showing some form of explanation).

To clarify how our set-up and assumptions differ from prior work, we overview work that we believe could be considered most similar to ours. We decompose our comparisons along a few dimensions: \textbf{Decision-support set-up:} Does the human make the final decision, or is it a different set-up? \textbf{Assumptions about decision-maker information:} What does prior work assume about access to a decision-maker when learning a policy?
\textbf{Evaluation:} Does prior work simulate humans? Does prior work run user studies?

\bigskip 
\noindent\citet{mozannar2020consistent} and \citet{hemmer2023human}:
\begin{itemize}[leftmargin=*]
    \item \textbf{Decision-support set-up:} This work's set-up can be considered a two-action setting of our formulation, where $\mathcal{A} = \{ \texttt{DEFER}, \texttt{MODEL}\}$. Extending the work of~\citet{madras2018predict}, this work proposes the learning to defer paradigm, where the decision-maker may not always make a final decision (i.e., sometimes the decision is deferred entirely to an algorithmic-based system). 
    In our set-up, deferring to a \texttt{MODEL} is equivalent to always adhering to a label-based form of support. 
    The human is always the final decision-maker in our work, which is representative of many decision-making set-ups in practice~\citep{lai2022human}, but not captured in this line of prior work. 
    \item \textbf{Assumptions about decision-maker information:} This work assumes oracle query access to the decision-maker, for whom they are learning a policy. 
    \item \textbf{Evaluation:} This work evaluates their approach using human simulations (no real human user studies). ~\citet{mozannar2020consistent} define synthetic experts in the following way: ``if the image belongs to the first $k$ classes the expert predicts perfectly, otherwise the expert predicts uniformly over all classes.''
\end{itemize}

\noindent\citet{gao2021human} and \citet{gao2023learning}:
\begin{itemize}[leftmargin=*]
    \item \textbf{Decision-support set-up:} This work defines two actions: $\mathcal{A} = \{ \texttt{DEFER}, \texttt{MODEL}\}$. They do not consider the, more practical assumption that the decision-maker will view a model prediction before making a decision themselves.
    Their formulation is similar to the above but they use offline bandits to learn a suitable policy.
    \item \textbf{Assumptions about decision-maker information:} \citet{gao2021human} assume that understanding decision-maker's expertise (at a population-level, not at a individual-level) can help learn better routing functions (i.e., defer only when appropriate).  \citet{gao2023learning} assume access to a decision history for each decision-maker.
    \item \textbf{Evaluation:} They run a human subject experiment to collect offline annotations, which can be used to learn when to defer to decision-makers. \citet{gao2023learning} goes further to personalize a deferral policy based on offline annotations for each decision-maker. 
\end{itemize}

\noindent\citet{bordt2022bandit}: 
\begin{itemize}[leftmargin=*]
    \item \textbf{Decision-support set-up:} This work's set-up can be considered a two-action setting of our formulation, where $\mathcal{A} = \{ \texttt{DEFER}, \texttt{SHOW}\}$; however, they are concerned with the learnability of such a set up. They do not devise algorithms for this setting, as they are only focused on its theoretical formulation.
    \item \textbf{Assumptions about decision-maker information:} They assume the decision-maker has access to information not contained in the input but  still important to the task. 
    \item \textbf{Evaluation:} This is a theory paper, containing neither computational nor human subject experiments.
\end{itemize}

\noindent\citet{noti2022learning}: 
\begin{itemize}[leftmargin=*]
    \item \textbf{Decision-support set-up:}  This work's set-up can be considered as a two-action setting of our formulation, where $\mathcal{A} = \{ \texttt{DEFER}, \texttt{SHOW}\}$. This is not an online algorithm and as such, the policy does not update.
    \item \textbf{Assumptions about decision-maker information:} They assume access to a dataset of human decisions and that all decision-makers are similar (i.e., they deploy one policy for all decision-makers). 
    \item \textbf{Evaluation:} This work is one of few that runs a user study to evaluate their (fixed) policy on unseen decision-makers.
\end{itemize}

\noindent\citet{babbar2022utility}: 
\begin{itemize}[leftmargin=*]
    \item \textbf{Decision-support set-up:}  This work considers the two-action setting of our formulation, where $\mathcal{A} = \{ \texttt{DEFER}, \texttt{CONFORMAL}\}$. Their policy is learned offline, is the same for all decision-makers, and is not updated in real-time based on decision-maker behavior.
\item \textbf{Assumptions about decision-maker information:} They use CIFAR-10H~\cite{peterson2019human} to learn a population-level deferral policy. This assumes that we have annotations for each decision-maker for every datapoint and assumes that all new decision-makers have the same expertise profiles as the population average.
    \item \textbf{Evaluation:} This work runs a user study to evaluate their (fixed) policy. They show the benefits of \texttt{DEFER+CONFORMAL} over \texttt{CONFORMAL} or \texttt{SHOW} alone.
\end{itemize}

\noindent\citet{wolczynski2022learning}: 
\begin{itemize}[leftmargin=*]
    \item \textbf{Decision-support set-up:}  This work considers two actions per our formulation, where $\mathcal{A} = \{ \texttt{DEFER}, \texttt{SHOW}\}$. They learn a rule-based policy offline for each decision-maker. 
\item \textbf{Assumptions about decision-maker information:} They simulate human behavior by considering explicit functions of how human expertise may vary in input space.
    \item \textbf{Evaluation:} While this work does consider the human to be the final decision-maker, they only validate their proposal in simulation, not on actual human subjects.
\end{itemize}

\noindent\citet{ma2023should}: 
\begin{itemize}[leftmargin=*]
    \item \textbf{Decision-support set-up:}  This work considers two actions per our formulation, where $\mathcal{A} = \{ \texttt{DEFER}, \texttt{SHOW}\}$. They learn a tree-based and rule-based policy offline for the population but permit one-off editing of these policies.
\item \textbf{Assumptions about decision-maker information:} They assume access to offline data to initialize their policies, which are then editable. They do not automatically learn online from user behavior over future interactions. 
    \item \textbf{Evaluation:} They run extensive human subject experiments to show the benefit of fixed policies, learned from a population and edited by each decision-maker. On models trained on simple tabular UCI datasets, they qualitatively study a decision-maker's trust and perception of assistance. 
\end{itemize}

\noindent\citet{buccinca2024towards}: 
\begin{itemize}[leftmargin=*]
    \item \textbf{Decision-support set-up:} 
    This work considers four actions per our formulation, where $\mathcal{A} = \{ \texttt{DEFER}, \texttt{EXPLAIN},\texttt{CLICK},\texttt{SHOW},\}$. They learn a reinforcement learning policy offline for each decision-maker. This requires having data for each action.
\item \textbf{Assumptions about decision-maker information:} They require a purely exploratory phase in their learning procedure: this prior data can then be used to learn the policy.
    \item \textbf{Evaluation:} They run human subject experiments on a toy task for assigning exercises to characters, picking between two classes. This differs from our realistic MMLU setting where optionality is higher as well.
\end{itemize}

We now list various forms of support that can be included in the action space of our problem formulation. The design of the action space is up to domain experts, who can decide not only which actions are feasible but also how much cost to assign to each form of support.

\begin{itemize}[leftmargin=*]
    \item \texttt{DEFER}: This form of support is equivalent to \textbf{no support}. Decision-makers are asked to make a decision without any assistance. The machine learning community has studied how to identify when to defer to a subset of examples to humans based on human strengths~\cite{bansal2021most,wilder2021learning} and/or model failures~\cite{chow1957optimum,geifman2017selective}. The premise of such an action would be to allow human decision-makers to be unaided and squarely placing decision liability on the individual.
    \item \texttt{SHOW}: In many settings, machine learning (ML) models are trained to do prediction tasks similar to the decision-making task prescribed to the human, or in the case of foundation models~\citep{foundationModels}, ML systems can be adapted to aid decision-making, even if the task was not specifically prescribed at train-time~\citep{yang2023foundation}. In essence, a machine learning model prediction, or associated generation (e.g., a code snippet~\citep{mozannar2022reading}) would be shown to aid an individual decision-maker. This has been shown to help improve decision-maker performance. 
    The following are variations of showing a model prediction to a decision-maker.
    \begin{itemize}
    \item \texttt{CONFORMAL}: For classification tasks, only displaying the most likely label may not lead to good performance due to various reasons, including uncertainty in the modeling procedure~\citep{vovk2005,bondi2021role}; however, such uncertainty can be communicated to decision-maker by showing a prediction set to experts~\citep{babbar2022utility}. Such a prediction set might be generated using conformal prediction, which guarantees the true label lies in the set with a user-specified error tolerance~\citep{bates2021distribution,straitouri2022provably}.
    \item \texttt{CONFIDENCE}: Instead of translating the uncertainty into a prediction set (or interval), one could simply show the confidence or uncertainty of the prediction, which may manifest as displaying probabilities, standard errors, or entropies~\cite{spiegelhalter_risk_2017,bhatt2021uncertainty}. The visualization mechanism used for displaying confidence may alter the decision-maker's performance~\citep{hullman2018pursuit,zhang2020effect}.
    \item \texttt{EXPLAIN}: In addition to providing a model prediction, many have considered showing an explanation of model behavior, examples of which include feature attribution~\citep{ribeiro2016should,buccinca2020proxy}, sample importance~\citep{kim2014bayesian,jeyakumar2020can}, counterfactual explanations~\cite{ustun2019actionable,antoran2020getting}, and natural language rationales~\cite{ehsan2018rationalization,camburu2018snli}. Displaying such explanations to end users has had mixed results on how decision-making performance is affected~\cite{chen2022interpretable,lai2022human}. Worryingly, in many settings, showing some types of explanations may lead to to over-reliance on models by giving the perception of competence~\cite{buccinca2020proxy,zerilli2022transparency, chen2023understanding}.
    \end{itemize}
    
    \item \texttt{CONSENSUS}: One can also depict forms of support that are independent of any model, for instance, presenting the belief of one or more humans. Belief distributions can be constructed by pooling over many different humans' ``votes'' for what a label ought to be, e.g.,~\cite{peterson2019human, imagenetReaLH, uma2022scaling, Uma_Fornaciari_Hovy_Paun_Plank_Poesio_2020, gordon2021disagreement, gordon2022jury}, or by eliciting distributions over the likely label directly from each individual human~\cite{softLabelElicitingLearning2022, collins2023human}. These consensus distributions permit the expression of uncertainty \textit{without any model}. 
    However, the elicitation of this form of support may be costly and humans may be fallible in the information they provide, e.g., due to direct labeling errors \citep{dawid1979maximum, augustin2017bayesian, whitehill2009whoseVote, wei2022learning}, or miscalibrated confidence~\cite{uncertainJudgments, collins2023human, lichtenstein1977calibration, kahneman1996reality}.

    \item \texttt{ADDITIONAL}: While much of this paper focused on support that provides decision-makers with label information, decision support may also entail acquiring or displaying additional contextual information (e.g., new features~\cite{bakker2021beyond}) or requesting previously unseen features, for instance, through additional medical diagnostics ~\cite{harrell1982evaluating,mylonakis2000report}. This flavor of support can be varied structurally, ranging from the results of a search query~\citep{webgpt} to hierarchical information like exposing the subsidiary ownership structure for multinational corporations~\cite{erramilli1996nationality}. Some pieces of additional information may require additional cost, or certification if pertaining to sensitive attributes. 
\end{itemize}

\paragraph{Prior work on multi-objective contextual bandits.} We summarize why some theoretical work on multi-objective contextual bandits cannot be directly applied to our problem formulation. In their study, ~\citet{tekin2018multi} addressed a contextual multi-armed bandit problem with two objectives, where one objective dominates the other. Their aim was to maximize the total reward in the non-dominant objective while ensuring that the dominant objective's total reward is also maximized. However, our specific case requires the minimization of the total expected loss of the support policy while ensuring that the expected accuracy of the decision-maker under the policy remains above a certain threshold. Therefore, the techniques presented by here cannot be directly applied to our scenario. \citet{turgay2018multi} investigated the multi-objective contextual bandit problem with similarity information. Their approach relies on the assumption that a Lipschitz condition holds for the set of feasible context-arm pairs concerning the expected rewards for all objectives and that the learning algorithm has knowledge of the corresponding distance function (see Assumption 1 of their paper).

~

\section{Additional Details on Problem Formulation}

The optimization problem in the standard setting, where the only objective relates to expected loss, can be written as:
\begin{align*}
\min_{\pi \in \Pi} L_h(\pi) ~=~& \min_{\pi \in \Pi} \mathbb{E}_{(x,y) \sim \gP} \big[\mathbb{E}_{\textsc{A}_i \sim \pi(x)}[\ell (y, h(x, \textsc{A}_i))]\big]  \\
~\stackrel{(a)}{=}~& \min_{\pi \in \Pi} \mathbb{E}_{(x,y) \sim \gP} \bigg[\sum_{i=1}^k{\pi(x)_{\textsc{A}_i} \cdot \ell (y, h(x, \textsc{A}_i))}\bigg]  \\
~\stackrel{(b)}{=}~& \min_{\pi \in \Pi}   \mathbb{E}_x \bigg[\sum_{i=1}^k{\pi(x)_{\textsc{A}_i} \cdot \mathbb{E}_{y|x}[\ell (y, h(x, \textsc{A}_i))]}\bigg] \\
~\stackrel{(c)}{=}~& \min_{\pi \in \Pi}   \mathbb{E}_x \bigg[\sum_{i=1}^k{\pi(x)_{\textsc{A}_i} \cdot r_{\textsc{A}_i}(x; h)}\bigg] , 
\end{align*}
where $(a)$ is due to the notation $\pi(x)_{\textsc{A}_i} := \mathbb{P}[\textsc{A}_i \sim \pi(x)]$, $(b)$ is due to the properties of expectation, and $(c)$ is due to the notation $r_{\textsc{A}_i}(x; h) := \mathbb{E}_{y|x}[\ell (y, h(x, \textsc{A}_i))]$. Then, by noting the fact that the expression $\mathbb{E}_x \bigg[\sum_{i=1}^k{\pi(x)_{\textsc{A}_i} \cdot r_{\textsc{A}_i}(x; h)}\bigg]$ can be optimized independently for each $x \in \mathcal{X}$, we can rewrite the above optimization problem as follows, for each $x \in \mathcal{X}$:
\[
\min_{\pi(x) \in \Delta(\mathcal{A})}   \sum_{i=1}^k{\pi(x)_{\textsc{A}_i} \cdot r_{\textsc{A}_i}(x; h)} ~=~ \min_{\textsc{A}_i \in \mathcal{A}} r_{\textsc{A}_i}(x; h) .
\]
Thus, an optimal policy for the above optimization problem is: $\pi^*(x) \in \argmin_{\textsc{A}_i \in \mathcal{A}} r_{\textsc{A}_i}(x; h)$ with random tie-breaking.

\subsection{Implementations of \texttt{Modiste}} 
\label{app:algo-context}

The human decision-making process with various forms of support can be effectively modeled as a stochastic contextual bandit problem. In this model, the diverse forms of support represent the available arms, and $\gX$ represents the context space. For our investigation, we leverage two simple and efficient techniques from the existing contextual bandit literature: LinUCB~\citep{li2010contextual} and KNN-UCB~\citep{guan2018nonparametric}. It's worth noting that while any contextual bandit algorithm could be employed, we chose these two simpler methods as they have demonstrated their effectiveness in empirically showcasing the utility of our online decision support policy learning framework.
We present specific implementations of the \texttt{Modiste} algorithm (see Algorithm~\ref{alg:DSP}) using LinUCB and KNN-UCB in Algorithms~\ref{alg:rpu-LinUCB}~and~\ref{alg:rpu-KNN}, respectively. 

\begin{algorithm*}[htb]
\caption{\texttt{Modiste} using LinUCB}
\label{alg:rpu-LinUCB}
\begin{algorithmic}[1]
    \STATE \textbf{Input:} 
    human decision-maker $h$; 
    UCB parameter $\alpha$
    \STATE \textbf{Initialization:} data buffer $\gD_0 = \{\}$; $\mathbf{A}_a = \mathbf{I}_{p \times p}$ and $\mathbf{b}_a = \mathbf{0}_{p \times 1}$ for all $a \in \mathcal{A}$; $\{ \theta_a = (\mathbf{A}_a)^{-1} \mathbf{b}_a: a \in \gA \}$; human prediction error values $\{ \widehat{r}_{a, 0} (x; h) = \ipp{\theta_a}{x} : x \in \gX, a \in \gA \}$; initial policy $\pi_1(x)_a = 1/\abs{\gA}$ for all $a \in \gA$
    \FOR{$t = 1, 2, \dots, T$}
    \STATE data point $(x_t, y_t) \in \gX \times \gY$ is drawn iid from $\gP$ (normalized s.t. $\norm{x_t}_2 \leq 1$)
    \STATE support $a_t \in \gA$ is selected using policy $\pi_t$
    \STATE human makes the prediction $\widetilde{y}_t$ based on $x_t$ and $a_t$
    \STATE human incurs the loss $\ell(y_t, \widetilde{y}_t)$
    \STATE update the buffer $\gD_t \gets \gD_{t-1} \cup \bcc{(x_t, a_t, \ell(y_t, \widetilde{y}_t))}$
    \STATE update the decision support policy: \vspace{-2mm}
    \begin{align*}
    \mathbf{A}_{a_t} ~\gets~& \mathbf{A}_{a_t} + x_t x_t^\top \\
    \mathbf{b}_{a_t} ~\gets~& \mathbf{b}_{a_t} + r_{t} x_t \\
    \theta_a ~\gets~& (\mathbf{A}_a)^{-1} \mathbf{b}_a \text{ for all } a \in \gA \\
    \widehat{r}_{a, t} (x; h) ~\gets~& \ipp{\theta_a}{x} \text{ for all } a \in \gA \tag{Step 1} \\
    \pi_{t+1}(x) ~\gets~& \argmin_{a \in \mathcal{A}} \widehat{r}_{a,t}(x; h)  - \alpha \cdot \sqrt{x^\top (\mathbf{A}_a)^{-1} x} \tag{Step 2}
    \end{align*}
    \ENDFOR{}
    \STATE \textbf{Output:} policy $\pi^\text{alg} \gets \pi_{T+1}$
  \end{algorithmic}
\end{algorithm*}

\begin{algorithm*}[htb]
\caption{\texttt{Modiste} using online KNN}
\label{alg:rpu-KNN}
\begin{algorithmic}[1]
    \STATE \textbf{Input:} 
    human decision-maker $h$; 
    warm-up steps $W$; number of neighbours $K$; exploration parameter $\gamma$
    \STATE \textbf{Initialization:} data buffer $\gD_0 = \{\}$; human prediction error values $\{ \widehat{r}_{a, 0} (x; h) = 0.5 : x \in \gX, a \in \gA \}$; initial policy $\pi_1(x)_a = 1/\abs{\gA}$ for all $a \in \gA$
    \FOR{$t = 1, 2, \dots, T$}
    \STATE data point $(x_t, y_t) \in \gX \times \gY$ is drawn iid from $\gP$ 
    \STATE support $a_t \in \gA$ is selected using policy $\pi_t$
    \STATE human makes the prediction $\widetilde{y}_t$ based on $x_t$ and $a_t$
    \STATE human incurs the loss $\ell(y_t, \widetilde{y}_t)$
    \STATE update the buffer $\gD_t \gets \gD_{t-1} \cup \bcc{(x_t, a_t, \ell(y_t, \widetilde{y}_t))}$
    \STATE update the decision support policy: \vspace{-2mm}
    \begin{align*}
    \gN(x) ~\gets~& \text{K neighbouring data points for } x \text{ in } \gD_t \\
    \gN_a(x) ~\gets~& \bcc{(x_i, a_i, \ell(y_i, \widetilde{y}_i)): (x_i, a_i, \ell(y_i, \widetilde{y}_i)) \in \gN(x) \text{ and } a_i = a} \quad \text{ for all } a \in \gA \\
    \widehat{r}_{a, t} (x; h) ~\gets~& \frac{1}{\abs{\gN_a(x)}} \cdot \sum_{(x_i, a_i, \ell(y_i, \widetilde{y}_i)) \in \gN_a(x)}{\ell(y_i, \widetilde{y}_i)} \quad \text{ for all } a \in \gA \text{ with } \abs{\gN_a(x)} > 0 \tag{Step 1} \\
    \pi_{\text{rand}}(x)_a ~\gets~& \frac{1}{\abs{\gA}} \quad \text{ for all } a \in \gA \\
    \pi_{\text{knn}}(x) ~\gets~& \argmin_{a \in \mathcal{A}} \widehat{r}_{a,t}(x; h)\\
    \pi_{t+1}(x) ~\gets~& \pi_{\text{rand}}(x) \quad \text{ if } t \leq W \tag{Step 2} \\
    \pi_{t+1}(x) ~\gets~& \gamma \cdot \pi_{\text{rand}}(x) + (1 - \gamma) \cdot \pi_{\text{knn}}(x) \quad \text{ if } t > W \tag{Step 2}
    \end{align*}
    \ENDFOR{}
    \STATE \textbf{Output:} policy $\pi^\text{alg} \gets \pi_{T+1}$
  \end{algorithmic}
\end{algorithm*}

\section{Additional Experimental Details}\label{appdx:exp_details}

\begin{figure*}[tb]
    \centering
    \begin{minipage}{0.33\textwidth}
        \centering
        \includegraphics[width=\textwidth]{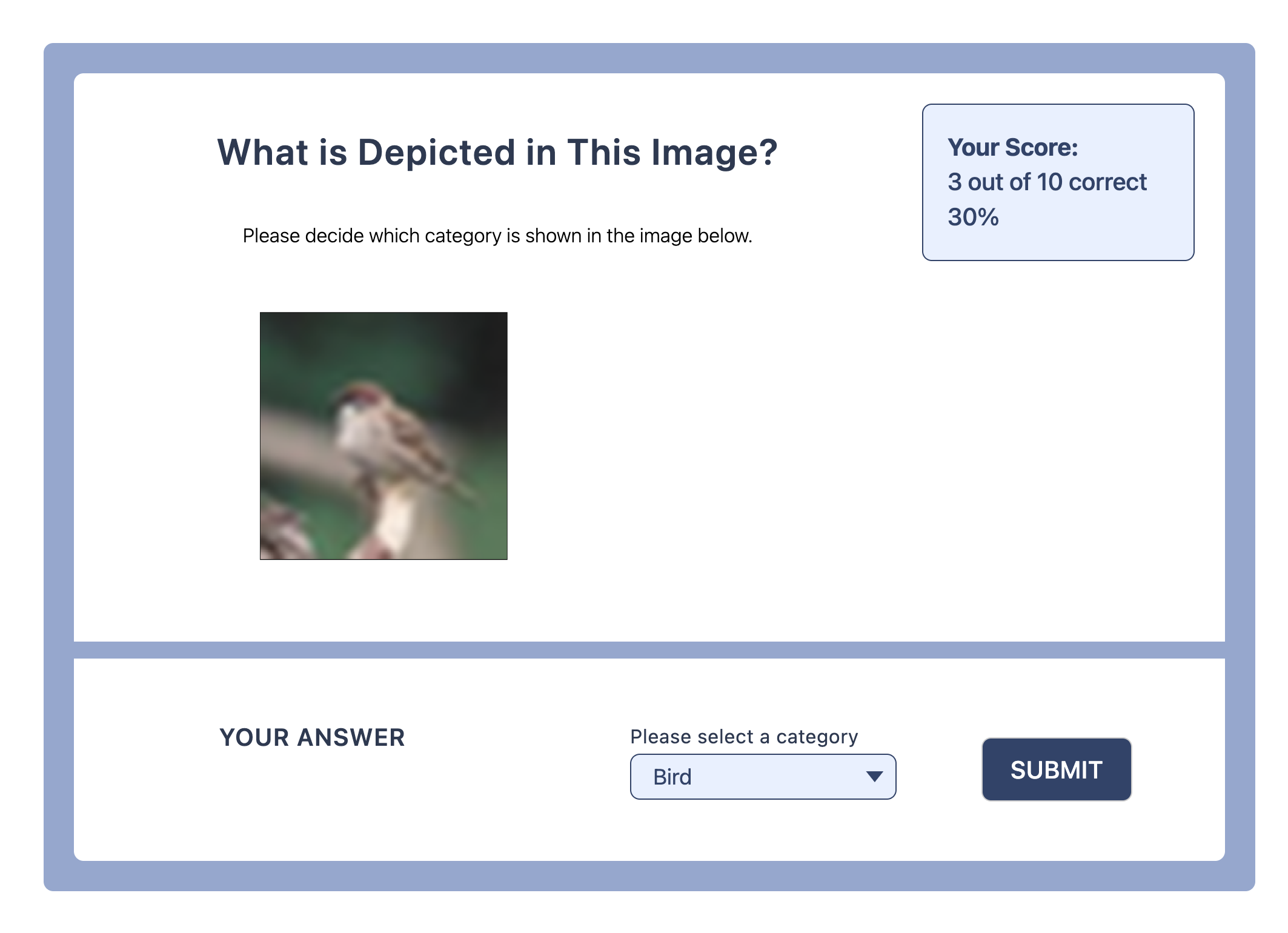} 
    \end{minipage}\hfill
    \begin{minipage}{0.33\textwidth}
        \centering
        \includegraphics[width=\textwidth]{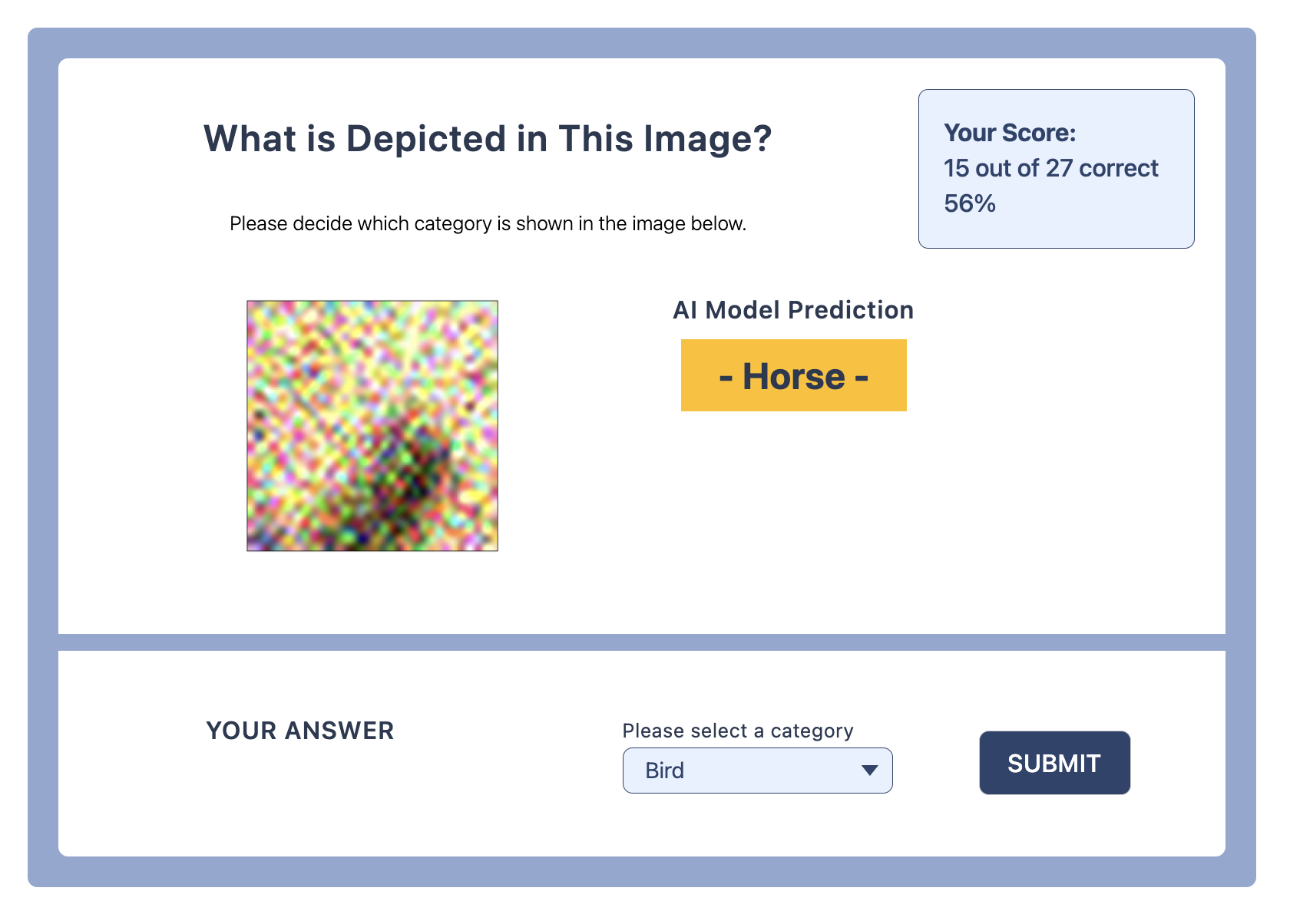} 
    \end{minipage}\hfill
    \begin{minipage}{0.33\textwidth}
        \centering
        \includegraphics[width=\textwidth]{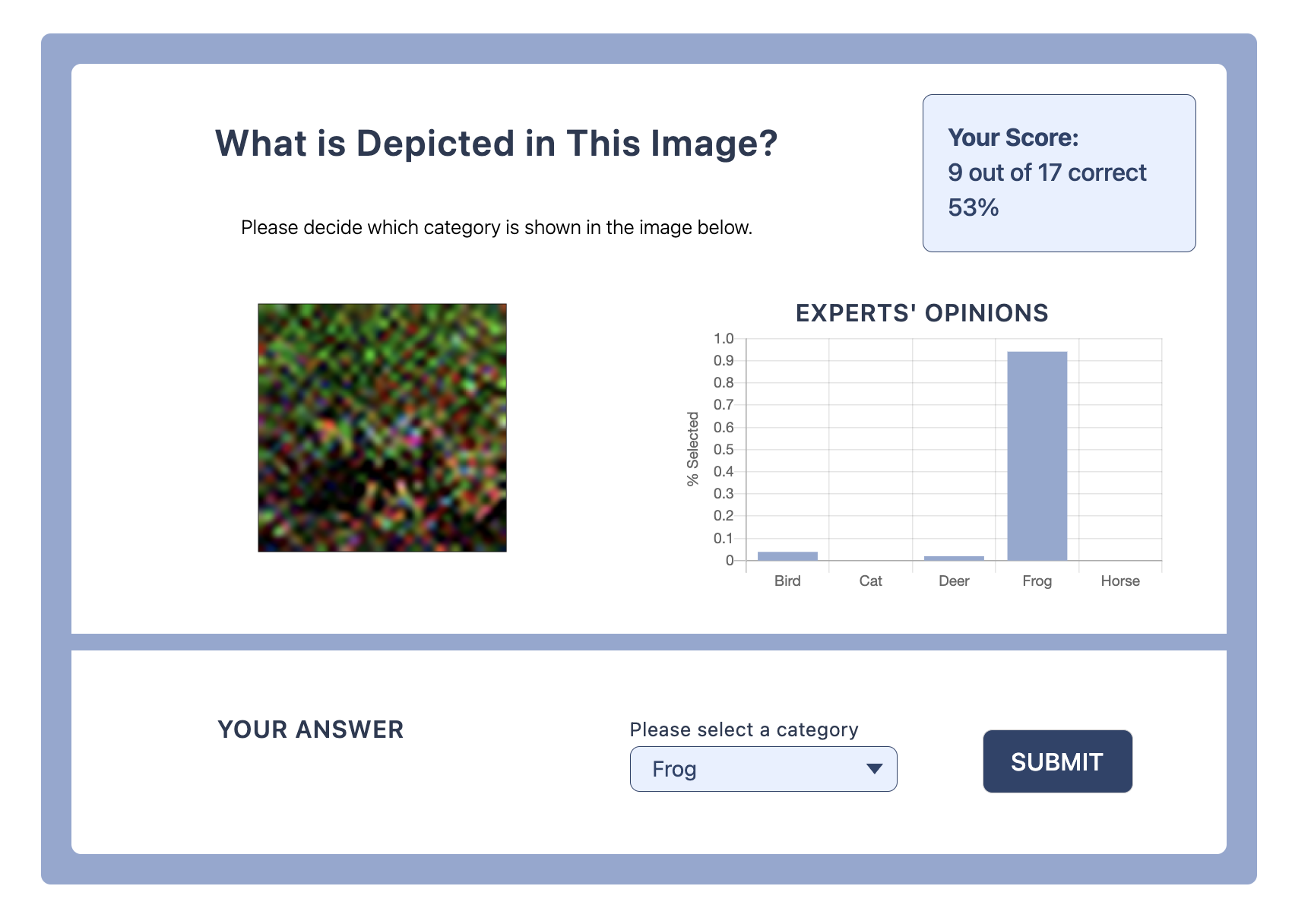} 
    \end{minipage}
    \hfill
    \begin{minipage}{0.42\textwidth}
        \centering
        \includegraphics[width=\textwidth]{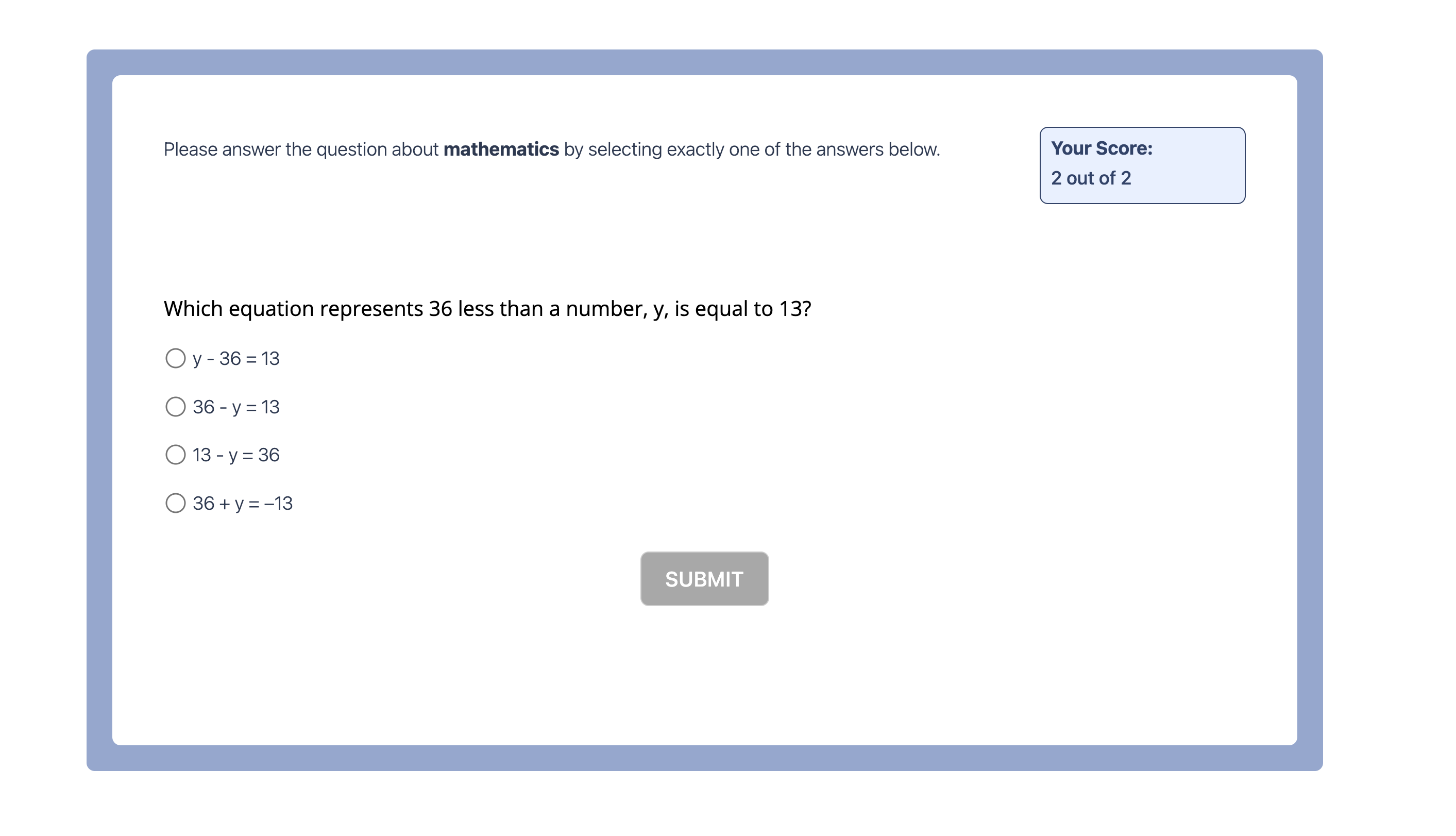} 
    \end{minipage}
    \begin{minipage}{0.42\textwidth}
        \centering
        \includegraphics[width=\textwidth]{figures/mmlu_llm.png} 
    \end{minipage}
    \hfill
    
    \caption{Examples of the \texttt{Modiste} interface for each form of support in CIFAR-$3A$ (Top) and MMLU-$2A$ (Bottom). From left to right, top row: \textsc{Human Alone}, \textsc{Model}, and \textsc{Consensus}; bottom row: \textsc{Human Alone} and \textsc{LLM}.
    }
        \label{fig:hse-interface-all}
\end{figure*}

\subsection{Datasets}

\paragraph{Computational Only.} Two tasks (Synthetic-$2A$ and CIFAR-$2A$) are used only in the computational experiments, so we describe them abstractly. Both have the same set-up, but have different underlying data generating distributions. Consider learning a decision support policy in a setting where there are \textit{two} forms of support available: $\mathcal{A} = \{\textsc{A}_1, \textsc{A}_2\}$, hence CIFAR-$2A$. 
We simulate a decision-maker's behavior under each $r_{\textsc{A}_i}$ across 3 classes for Synthetic/CIFAR. We instantiate the $m$ decision-makers in the population data using human simulations, where we randomly sample $r_{\textsc{A}_i}(x; h)$ from a distribution for each $h$ in the population.

\begin{figure}[htb]
     \centering
     \includegraphics[width=0.8\linewidth]{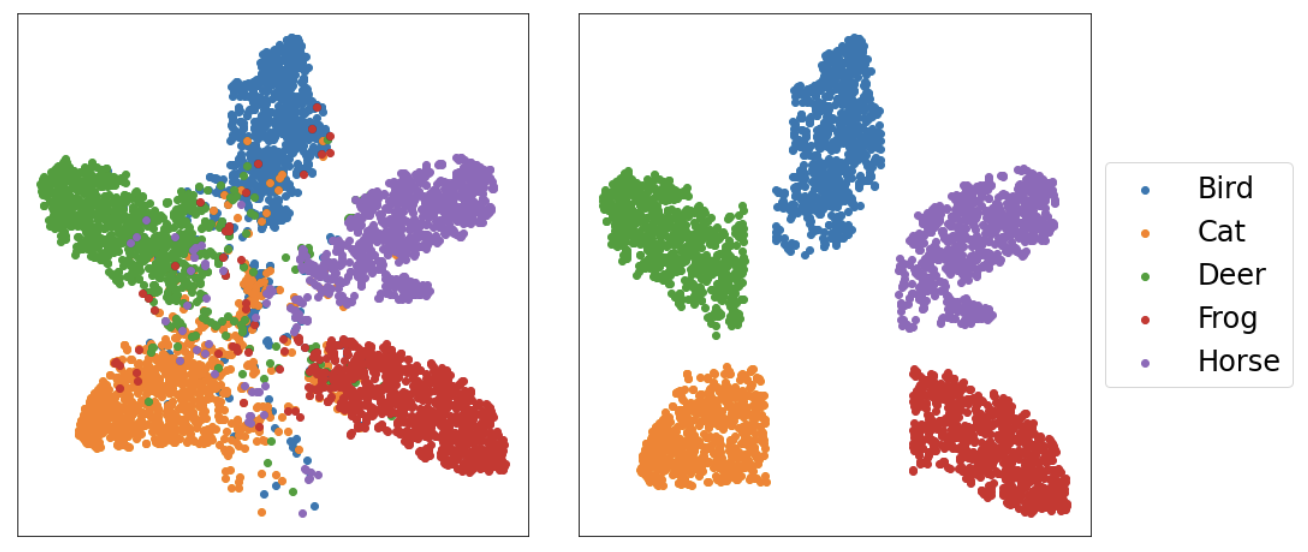}
     \caption{We depict the latent space of the CIFAR subset used. Embeddings without filtering (left) and after filtering out points that violate separability of classes (right). We use the embeddings on the right for our Synthetic-$2A$ task and the left for the natural CIFAR tasks.}
    \label{fig:separable_cifar}
\end{figure}

\paragraph{CIFAR.} To explore the performance of our algorithms in both separable and non-separable settings, we construct two subvariants of the CIFAR-10~\citep{krizhevsky2009learning} image dataset. As depicted in Figure \ref{fig:separable_cifar}, we subsample from the CIFAR embeddings to create linearly separable classes. Embeddings are constructed by running t-SNE on the 512-dimensional latent codes extracted by the penultimate layer of a variant of the VGG architecture (VGG-11) \citep{he2016deep} trained on the animal class subset of CIFAR; the model attained an accuracy of 89.5\%. 
We filter the original 10 CIFAR classes to only include those involving animals (Birds, Cats, Deers, Dogs, Frogs, and Horses). We consider at most 5 such classes in our experiments (dropping Dog).

\subsubsection{MMLU}

We consider questions from four topics of MMLU (elementary mathematics, high school biology, high school computer science, and US foreign policy). Topic names match those proposed in \citep{hendrycks2020measuring}. We select topics to cover span an array of disciplines across the sciences and humanities, as discussed in Appendix \ref{appdx:user_study}. We use questions in the MMLU test set for each topic that are at most 150 characters long. We limit the length of questions to facilitate readability in our human user studies and wanted to maximize parity between our computational and human experiments by considering the same set of questions across both set-ups. This yields 264, 197, 98, and 47 questions for the elementary mathematics, high school biology, US foreign policy, and high school computer science topics, respectively. We emphasize that in CIFAR, our forms of support operate in label space -- in contrast, with MMLU, support selection is in \textit{covariate} (topic) space -- further highlighting the flexibility of our adaptive support paradigm.

We leverage OpenAI's LLM-based embedding model (\texttt{text-embedding-ada-003}), to produce embeddings over the question prompts. Embedding vectors extracted via OpenAI's API are by default length 1536; we apply t-SNE, like in the CIFAR tasks, to compress the embeddings into two-dimensional latent codes per example. Nicely, these latents are already separable (see Figure \ref{fig:hse-mmlu-topology}; no post-filtering is applied to ensure separability).

\subsection{CIFAR Task Set-up}\label{appdx:corruption}

We consider a 5-class subset of CIFAR (Bird, Cat, Deer, Frog, and Horse), as discussed in Appendix \ref{appdx:comp_exp}\footnote{We explored a 3 class variant for HSEs to directly match the computational experiments; however, we realized that participants were able to figure out that which classes were impoverished, raising the base rate of correctly categorizing such images. 
Such behavior again highlights the need to carefully consider real human behavior in adaptive decision support systems.}. We instantiate two forms of support: 1) a simulated AI model which provides a prediction for the image class, and 2) a consensus response derived from real humans, derived from the approximately 50 human annotators from CIFAR-10H \citep{peterson2019human, battleday2020capturing}, presented as a distribution over the 5 classes. 

We treat the original CIFAR-10 test set labels as the ``true'' labels and only include images for which the original CIFAR-10 label matches the label deemed most likely from the CIFAR-10H annotators (discrepancy was a rare occurrence, only 1.1\% of the 5000 images considered). We sample a pool of 300 such images, and construct three different batches of 100 images. Participants are assigned to one batch of 100 images. Images per batch are sampled in a class-balanced fashion (i.e., 20 images for each of the 5 categories). Images are shuffled for each participant. The same latent codes ($z_t$) used in the computational experiments are employed for the user studies.

\subsubsection{Corruptions and Support Design}

We want to study whether our algorithms can properly learn which forms of decision support are actually needed by real humans. We therefore need to ensure that the task is sufficiently rich such that humans do \textit{need} support (and the existing forms of support which can be of value). To mimic such a setting in CIFAR, we deliberately corrupt images of all classes except one (i.e., Birds\footnote{We further disambiguate the non-corrupted bird class by upsampling to 160x160 images via Lanczos-upsampling following \citep{peterson2019human, battleday2020capturing, softLabelElicitingLearning2022} before presenting them to participants.}) when presenting images to the user. Corruptions are formed via a composition of natural adversarial transformations proposed in~\citet{hendrycks2020measuring}, specifically shot noise ontop of glass blur. This enables us to check that our algorithms are able to properly recommend support for all other classes, as a human will be unable to decide on image category unassisted. Example corrupted images can be seen in Figure \ref{fig:hse-interface}. 

To ensure that the available forms of support have different regions of strength, we enforce that each form of support is only good at two of the five classes. In the case of the simulated AI model, we return the ``true'' class whenever the image is a Deer or Cat classes, and return one of the incorrect classes for all other images. For the consensus labels, we use the CIFAR-10H distribution derived from 50 annotators when the image is a Horse or Frog. As the CIFAR-10H labels were originally collected over all 10 CIFAR classes, sometimes an annotation was endorsed for a class outside of the 5 we consider; in that case, we discard the annotations assigned to those classes and renormalize the remaining distribution (on average, $< 2\%$ of the original labeling mass was discarded per image). For all other classes, we intend the consensus to be an unhelpful form of support, as such, for images in the Bird, Deer, and Horse class, we return a uniform distribution (i.e., providing no information to the participant).

\subsection{MMLU Task Set-up}\label{appdx:mmlu}

Participants respond to 60 multiple choice questions, which were balanced by topic (15 questions per topic). Participants are assigned to one of three possible batches of 60 such questions. Order is shuffled. Participants are informed of the topic associated with each question (e.g., that the question was about biology). We implement a 10 second delay between when the question is presented and when the participant is allowed to submit their response to encourage participants to try each problem in earnest.

\subsubsection{Topic Selection}

We ran several pilot studies with \texttt{Modiste} to determine people's base performance on a subset of MMLU topics. We intended to select a diverse set of topics such that it was unlikely any one participant would excel at all topics -- as many of the topics were specialized and challenging \citep{hendrycks2020measuring} -- but also varied enough that participants may be strong in at least one area. We found that a large number of participants achieved reasonable performance on elementary mathematics, and that a sufficient number of participants also excelled at questions in the high school biology, US foreign policy, and high school computer science topics -- though usually not all together. 

We also factored in \texttt{InstructGPT-3.5}'s performance while deciding on the topics. To check whether our adaptive decision support algorithms are effective at learning good policies -- like with CIFAR -- it is helpful to have support available that is effective, should someone be unable to answer adequately alone. As a result, we looked more sympathetically on categories where model accuracy was already high.

However, the real-world is not so perfect: there may not be high-performing support available when a human struggles to make a decision. As such, we also deliberately forced down the accuracy of the LLM form of support in the mathematics topic. While we expect that most participants would be able to solve elementary mathematics problems without the aid of the LLM support, should they be unable to, the LLM was not able to help them. We leave further impovershing studies which mimic real-world support settings for future work.  

\subsubsection{Question Selection and Model Accuracy}

The resulting accuracy of the LLM form of support on the questions shown to humans per topic is 29\% for elementary mathematics, 89\% for high school biology, 87\% for US foreign policy, and 91\% for high school computer science. We upweighted selection of foreign policy questions that participants in our pilot had gotten correct when constructing the question subset, as we wanted to ensure that, should a participant have foreign policy experience, they would be able to answer them.

\subsection{Compute Resources}

All computational experiments were run using CPUs (either on local machines or Google Colab), with the exception of training the VGG used for CIFAR embeddings. Here, we accessed a group compute cluster and trained the model for roughly five hours on a Nvidia A100-SXM-80GB GPU.

\section{Additional Computational Experiments}\label{appdx:comp_exp}

\begin{figure*}[htb!]
    \centering
    \begin{minipage}{0.8\textwidth}
        \centering
        \includegraphics[width=\textwidth]{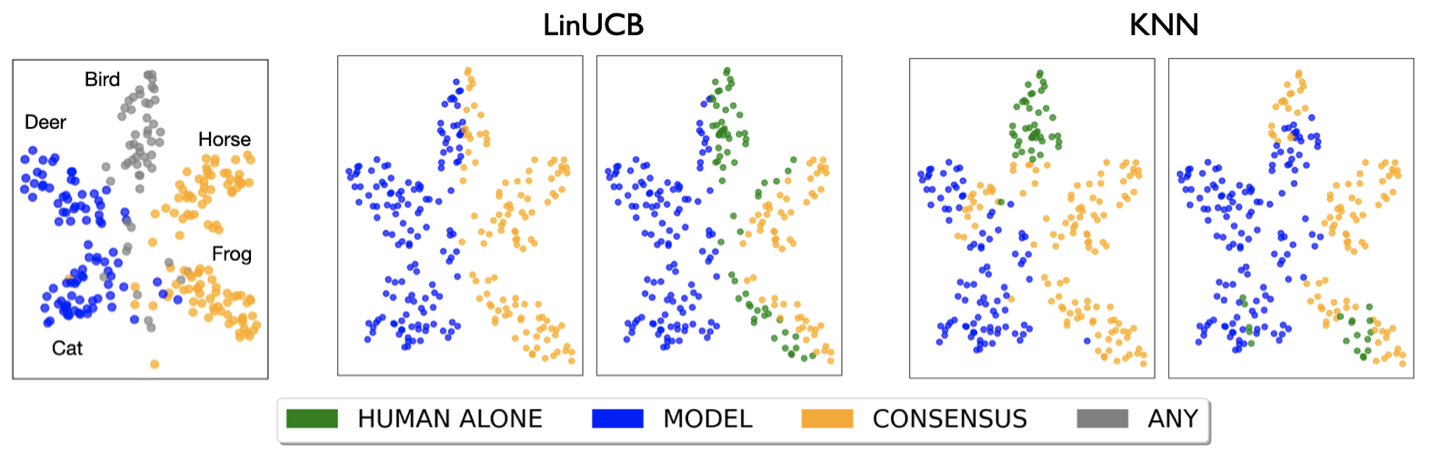} 
    \end{minipage}
    \caption{
    Snapshots of the learned decision support policies computed at the end of $100$ interactions with randomly sampled users for CIFAR-$3A$.
    The forms of support are colored in t-SNE embedding space. 
    Both LinUCB and KNN generally identify the optimal form of support for a given input. 
    All participant plots are included in Figure \ref{fig:hse-topology-more-cifar}.
    }

    \label{fig:hse-cifar-topology}
\end{figure*}

\begin{table*}[htb]
\centering
\footnotesize
\caption{Vary exploration parameters ($\alpha$ for LinUCB and $\gamma$ for KNN). We find that LinUCB generally performs better when exploration increases and KNN generally performs better when exploration decreases.
}
\resizebox{\textwidth}{!}{\begin{tabular}{c|c|c|c|c}
\hline
 & \small Algorithm & \begin{tabular}[c]{@{}c@{}}$r_{\textsc{A}_1} = [0.7, 0.1, 0.7]$\\ $r_{\textsc{A}_2} = [0.1, 0.1, 0.1]$\end{tabular} & \begin{tabular}[c]{@{}c@{}}$r_{\textsc{A}_1} = [0.7, 0.1, 0.7]$\\ $r_{\textsc{A}_2} = [0.1, 0.7, 0.1]$\end{tabular} & \begin{tabular}[c]{@{}c@{}}$r_{\textsc{A}_1} = [0.7, 0.1, 0.7]$\\ $r_{\textsc{A}_2} = [0.7, 0.7, 0.1]$\end{tabular} \\ \hline
\multirow{4}{*}{Synthetic-$2A$} & \small\texttt{Modiste}-LinUCB $\alpha=0.1$ & $0.14 \pm 0.04$ & $0.19 \pm 0.09$ & $0.34 \pm 0.10$ \\ \cline{2-5} 
 & \small\texttt{Modiste}-LinUCB $\alpha=10$ & $0.13 \pm 0.04$ & $0.12 \pm 0.03$ & $0.30 \pm 0.05$ \\ \cline{2-5} 
 & \small\texttt{Modiste}-KNN $\gamma=0.01$ & $0.12 \pm 0.04$ & $0.19 \pm 0.08$ & $0.33 \pm 0.04$ \\ \cline{2-5} 
 & \small\texttt{Modiste}-KNN $\gamma=0.2$ & $0.16 \pm 0.05$ & $0.21 \pm 0.06$ & $0.34 \pm 0.06$ \\ \hline
\multirow{4}{*}{CIFAR-$2A$} & \small\texttt{Modiste}-LinUCB $\alpha=0.1$ & $0.15 \pm 0.08$ & $0.14 \pm 0.05$ & $0.31 \pm 0.05$ \\ \cline{2-5} 
 & \small\texttt{Modiste}-LinUCB $\alpha=10$ & $0.14 \pm 0.04$ & $0.18 \pm 0.04$ & $0.32 \pm 0.05$ \\ \cline{2-5} 
 & \small\texttt{Modiste}-KNN $\gamma=0.01$ & $0.12 \pm 0.03$ & $0.14 \pm 0.06$ & $0.35 \pm 0.06$ \\ \cline{2-5} 
 & \small\texttt{Modiste}-KNN $\gamma=0.2$ & $0.17 \pm 0.05$ & $0.20 \pm 0.09$ & $0.35 \pm 0.06$ \\ \hline
\end{tabular}}
\ \\
\ \\
\ \\
\resizebox{\textwidth}{!}{\begin{tabular}{c|c|c|c|c}
\hline
                           &                                    Algorithm                                    & \begin{tabular}[c]{@{}c@{}}$r_{\textsc{A}_1} = [0.8, 0.8, 0.1, 0.1]$\\ $r_{\textsc{A}_2} = [0.2, 0.1, 0.1, 0.2]$\end{tabular} & \begin{tabular}[c]{@{}c@{}}$r_{\textsc{A}_1} = [0.2, 0.8, 0.1, 0.1]$\\ $r_{\textsc{A}_2} = [0.8, 0.1, 0.1, 0.2]$\end{tabular} & \begin{tabular}[c]{@{}c@{}}$r_{\textsc{A}_1} = [0.2, 0.5, 0.5, 0.8]$\\ $r_{\textsc{A}_2} = [0.8, 0.1, 0.1, 0.8]$\end{tabular} \\ \hline
\multirow{4}{*}{MMLU-$2A$} & \small\texttt{Modiste}-LinUCB $\alpha=0.1$ & $0.26 \pm 0.06$                                                                                                               & $0.22 \pm 0.09$                                                                                                               & $0.47 \pm 0.08$                                                                                                               \\ \cline{2-5} 
                           & \small\texttt{Modiste}-LinUCB $\alpha=10$  & $0.15 \pm 0.05$                                                                                                               & $0.18 \pm 0.07$                                                                                                               & $0.40 \pm 0.06$                                                                                                               \\ \cline{2-5} 
                           & \small\texttt{Modiste}-KNN $\gamma=0.01$   & $0.16 \pm 0.05$                                                                                                               & $0.15 \pm 0.06$                                                                                                               & $0.39 \pm 0.06$                                                                                                               \\ \cline{2-5} 
                           & \small\texttt{Modiste}-KNN $\gamma=0.2$    & $0.15 \pm 0.03$                                                                                                               & $0.12 \pm 0.03$                                                                                                               & $0.38 \pm 0.05$                                                                                                               \\ \hline
\end{tabular}}
\label{tab:exploration}
\end{table*}



\paragraph{Varying KNN Parameters.} While LinUCB does not require identifying additional parameters, KNN has two: $K$ which is the number of nearest neighbors to select when estimating the risk of a form of support and $W$ which is the length of the warm-up period. In Table~\ref{tab:knn_param}, we show that as long as $K$ is reasonably sized (i.e. $K > 3$), the performance of KNN does not vary too much across datasets.

\begin{table*}[htb]
\centering
\caption{We vary the $K$ and $W$ parameters used to instantiate KNN across three datasets and report the expected risk $L_h(\pi)$ (lower is better) across 5 runs. }
\begin{tabular}{c|c|c|c|c}
\hline
 & $K=3$, $W=10$ & \multicolumn{1}{l|}{$K=5$, $W=10$} & $K=5$, $W=25$ & \multicolumn{1}{l}{$K=8$, $W=40$} \\ \hline
Synthetic-$2A$ & $0.20 \pm 0.11$ & $0.15 \pm 0.05$ & $0.14 \pm 0.05$ & 0.12 $\pm$ 0.04 \\ \hline
CIFAR-$2A$ & $0.25 \pm 0.15$ & $0.14 \pm 0.04$ & $0.17 \pm 0.04$ & $0.14 \pm 0.04$ \\ \hline
MMLU-$2A$ & $0.18 \pm 0.07$ & $0.14 \pm 0.04$ & $0.15 \pm 0.03$ & $0.15 \pm 0.04$ \\ \hline
\end{tabular}
\label{tab:knn_param}
\end{table*}

\begin{table*}[htb]
\centering
\caption{On the CIFAR-$2A$ dataset, we explore the effect of varying the embedding size for KNN, with $K = 8$, $W = 25$ (Top), and LinUCB (Bottom). We report the expected risk $L_h(\pi)$ (lower is better) averaged over 5 runs. Recall that in our main paper experiments we used two-dimensional t-SNE embeddings, not these model embeddings, for ease of visualization.}
\begin{tabular}{c|cc|cc}
\hline
\begin{tabular}[c]{@{}c@{}}Embedding \\ Size\end{tabular} & \multicolumn{2}{c|}{\begin{tabular}[c]{@{}c@{}}$r_{\textsc{A}_1} = [0.7, 0.1, 0.7]$\\ $r_{\textsc{A}_2} = [0.1, 0.7, 0.1]$\end{tabular}} & \multicolumn{2}{c}{\begin{tabular}[c]{@{}c@{}}$r_{\textsc{A}_1} = [0.7, 0.1, 0.7]$\\ $r_{\textsc{A}_2} = [0.7, 0.7, 0.1]$\end{tabular}} \\ \cline{2-5} 
                                                          & \multicolumn{1}{c|}{$T = 100$}                                                & $T = 200$                                                & \multicolumn{1}{c|}{$T = 100$}                                                & $T = 200$                                               \\ \hline
2                                                         & \multicolumn{1}{c|}{$0.18 \pm 0.08$}                                          & $0.14 \pm 0.04$                                          & \multicolumn{1}{c|}{$0.32 \pm 0.05$}                                          & $0.33 \pm 0.06$                                         \\ \hline
4                                                         & \multicolumn{1}{c|}{$0.18 \pm 0.04$}                                          & $0.16 \pm 0.04$                                          & \multicolumn{1}{c|}{$0.33 \pm 0.05$}                                          & $0.33 \pm 0.05$                                         \\ \hline
8                                                         & \multicolumn{1}{c|}{$0.15 \pm 0.05$}                                          & $0.15 \pm 0.03$                                          & \multicolumn{1}{c|}{$0.32 \pm 0.05$}                                          & $0.33 \pm 0.05$                                         \\ \hline
512                                                       & \multicolumn{1}{c|}{$0.15 \pm 0.05$}                                          & $0.15 \pm 0.03$                                          & \multicolumn{1}{c|}{$0.35 \pm 0.05$}                                          & $0.35 \pm 0.05$                                         \\ \hline
\end{tabular}
\ \\
\ \\
\ \\
\begin{tabular}{c|cc|cc}
\hline
\begin{tabular}[c]{@{}c@{}}Embedding \\ Size\end{tabular} & \multicolumn{2}{c|}{\begin{tabular}[c]{@{}c@{}}$r_{\textsc{A}_1} = [0.7, 0.1, 0.7]$\\ $r_{\textsc{A}_2} = [0.1, 0.7, 0.1]$\end{tabular}} & \multicolumn{2}{c}{\begin{tabular}[c]{@{}c@{}}$r_{\textsc{A}_1} = [0.7, 0.1, 0.7]$\\ $r_{\textsc{A}_2} = [0.7, 0.7, 0.1]$\end{tabular}} \\ \cline{2-5} 
                                                          & \multicolumn{1}{c|}{$T = 100$}                                                & $T = 200$                                                & \multicolumn{1}{c|}{$T = 100$}                                                & $T = 200$                                               \\ \hline
2                                                         & \multicolumn{1}{c|}{$0.31 \pm 0.05$}                                          & $0.32 \pm 0.04$                                          & \multicolumn{1}{c|}{$0.49 \pm 0.05$}                                          & $0.49 \pm 0.05$                                         \\ \hline
4                                                         & \multicolumn{1}{c|}{$0.14 \pm 0.04$}                                          & $0.14 \pm 0.04$                                          & \multicolumn{1}{c|}{$0.32 \pm 0.04$}                                          & $0.31 \pm 0.50$                                         \\ \hline
8                                                         & \multicolumn{1}{c|}{$0.13 \pm 0.04$}                                          & $0.13 \pm 0.03$                                          & \multicolumn{1}{c|}{$0.31 \pm 0.03$}                                          & $0.31 \pm 0.04$                                         \\ \hline
512                                                       & \multicolumn{1}{c|}{$0.17 \pm 0.03$}                                          & $0.16 \pm 0.04$                                          & \multicolumn{1}{c|}{$0.34 \pm 0.04$}                                          & $0.33 \pm 0.06$                                         \\ \hline
\end{tabular}
\label{tab:embed}
\end{table*}

\paragraph{Varying Embedding Size.} 
In the main text, we run computational experiments using a two-dimension t-SNE embedding. In Table~\ref{tab:embed} for CIFAR-$2A$, we consider how \texttt{Modiste} would behave when we vary dimensionality of $\mathcal{X}$.  
We extract higher dimensional embeddings from animal-class trained VGGs (see above) with varied penultimate layer widths, where width matches embedding size. While we opt for t-SNE embeddings, in the main paper, to provide a clearer visualizations, we find that strong performance holds even for high-dimensional embeddings. We expect that even larger embedding dimensions will permit \texttt{Modiste} to work in a variety of real-world contexts, but may come at the added cost of an increase in the number of required interactions, if we learn with no prior assumptions on the decision-maker's expertise profile.

\begin{figure*}[htb]
    \centering
    \begin{minipage}{0.33\textwidth}
        \centering
        \includegraphics[width=\textwidth]{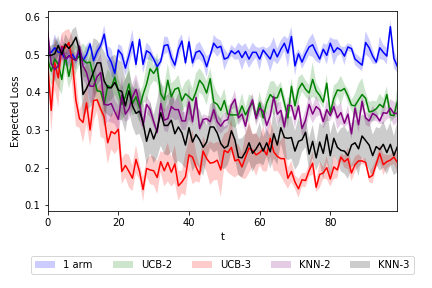} 
        \caption*{
    $r_{\textsc{A}_1} = [0.1, 0.7, 0.7]$\\ 
$r_{\textsc{A}_2} = [0.7, 0.1, 0.7]$\\
$r_{\textsc{A}_3} = [0.7, 0.7, 0.1]$
}
    \end{minipage}\hfill
    \begin{minipage}{0.33\textwidth}
        \centering
        \includegraphics[width=\textwidth]{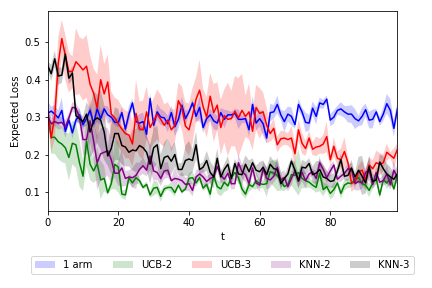} 
        \caption*{
    $r_{\textsc{A}_1} = [0.1, 0.7, 0.1]$\\ 
$r_{\textsc{A}_2} = [0.7, 0.1, 0.1]$\\
$r_{\textsc{A}_3} = [0.7, 0.7, 0.7]$
}
    \end{minipage}
    \begin{minipage}{0.33\textwidth}
        \centering
        \includegraphics[width=\textwidth]{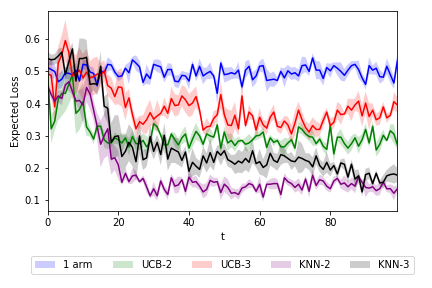} 
        \caption*{
            $r_{\textsc{A}_1} = [0.1, 0.7, 0.7]$\\ 
            $r_{\textsc{A}_2} = [0.7, 0.1, 0.1]$\\
            $r_{\textsc{A}_3} = [0.7, 0.7, 0.7]$
        }
    \end{minipage}
        \caption{Loss plots over time for human simulations (denoted by $r_{\textsc{A}_i}$) in the CIFAR-$3A$ setting.
    }
    \label{fig:arm_loss_plots}

\end{figure*}

\begin{table*}[htb]
\centering
\footnotesize
\caption{We evaluate $|\mathcal{A}| > 3$ via $L_h(\pi_t)$ for $t=50,100,250$ for the same setup as Table~\ref{tab:generalsetting}, but with more arms. We find that \texttt{Modiste} works well over time. The success of LinUCB is highly dependent on the embedding space and its geometry.
}
\begin{tabular}{cc|c|c|c}
\hline
 & Algorithm     & \begin{tabular}[c]{@{}c@{}}$T=50$\end{tabular} & \begin{tabular}[c]{@{}c@{}}$T=100$\end{tabular} & \begin{tabular}[c]{@{}c@{}}$T=250$\end{tabular} \\ \hline
\multicolumn{1}{c|}{\multirow{2}{*}{CIFAR-$5A$}}
& \texttt{Modiste}-LinUCB &   $0.59 \pm 0.10$   &      $0.29 \pm 0.17$                 &          $0.34 \pm 0.10$       \\ \cline{2-5} 
\multicolumn{1}{c|}{}                          
& \texttt{Modiste}-KNN   &   $0.19 \pm 0.15$   &      $0.14 \pm 0.17$                 &          $0.11 \pm 0.15$   \\ \hline
\multicolumn{1}{c|}{\multirow{2}{*}{MMLU-$4A$}} 
& \texttt{Modiste}-LinUCB &   $0.13 \pm 0.10$   &      $0.10 \pm 0.07$                 &          $0.17 \pm 0.09$           \\ \cline{2-5} 
\multicolumn{1}{c|}{}                          
& \texttt{Modiste}-KNN    &   $0.11 \pm 0.12$   &      $0.08 \pm 0.12$                 &          $0.02 \pm 0.04$         \\ \hline
\end{tabular}
\label{tab:multiarm}
\end{table*}

\begin{figure*}[htb]
     \centering
     \includegraphics[width=1.0\linewidth]{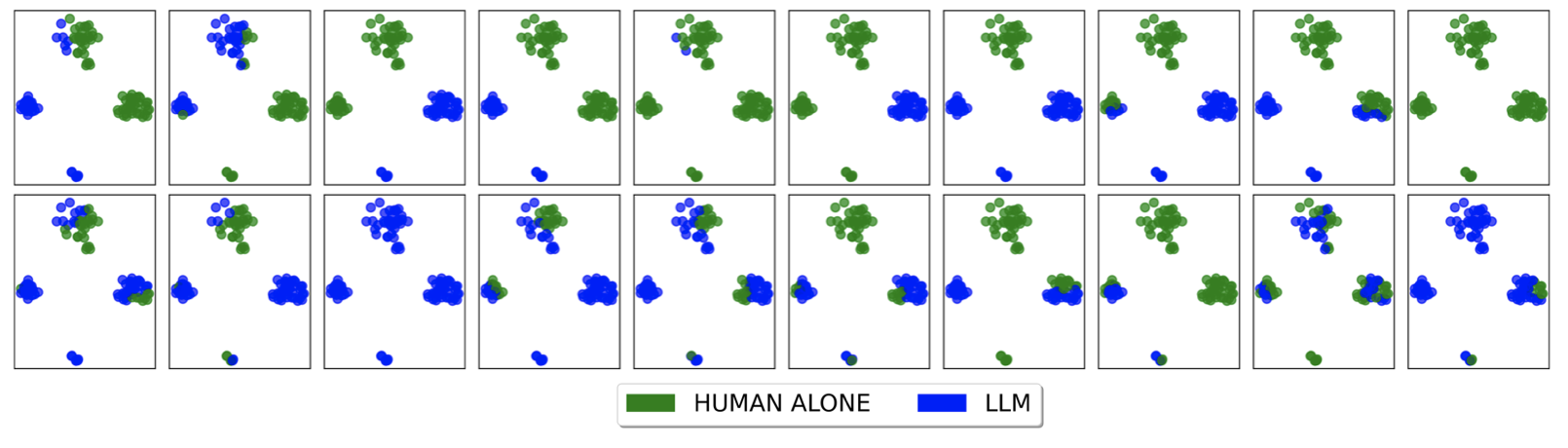}
     \caption{Snapshots of the learned decision support policies computed at the end of $60$ interactions with randomly sampled users for MMLU-$2A$ with LinUCB policies displayed atop KNN ones.
    The forms of support are colored in t-SNE embedding space.
    Topics correspond to clusters, clockwise from the top: math, biology, computer science, foreign policy.}
    \label{fig:hse-mmlu-topology_app}
\end{figure*}

\paragraph{$|\mathcal{A}|>3$ Experiments.} 
We now demonstrate how \texttt{Modiste} behaves when we increase the number of forms of support available to a decision-maker. Using the same set-up and algorithm-specific hyperparameters as Table~\ref{tab:generalsetting}, we convert the CIFAR setting to five forms of support (one for each class), resulting in CIFAR-$5A$, and convert the MMLU setting to four forms of support (one for each topic) resulting in MMLU-$4A$. In Table~\ref{tab:multiarm}, we report $L_h(\pi)$ at three time steps: $t = 50, 100, 250$. For CIFAR, we find that KNN significantly outperforms  given more interactions, which is expected when we increase  $|\mathcal{A}|$. We believe that LinUCB struggles as the t-SNE embedding space is likely not rich enough to capture the intricacies of all five forms of support. Given a larger embedding space, we expect that LinUCB would excel when the number of forms of support is increased. For MMLU, both LinUCB and KNN perform well by the 250th interaction: KNN is very good at lowering $L_h(\pi)$ on this task. In all, this experiment assures us that careful design of decision support can permit us to use \texttt{Modiste} when we have a larger set of potential forms of support. 

While we include experiments with larger $|\mathcal{A}|$ here, we note that higher set sizes may also be problematic for two reasons: 1) decision-makers may struggle under many forms of support~\citep{kalis2013we}, and 2) as size increases, we find that the
number of interactions required to learn an accurate personalized policy also increases.

\paragraph{Sensitivity to choice of $T$.}
In Figure~\ref{fig:arm_loss_plots}, we consider a setting of learning a policy with three forms of support: $\mathcal{A} = \{\textsc{A}_1, \textsc{A}_2, \textsc{A}_3\}$. 
This is an analogous setup to what we do in our human subject experiments in Section~\ref{sec:user_study}.
We indicate the simulated human expertise profiles in the caption. We show how performance on a random sample of 100 points from a held-out set varies with time while learning a decision support policy. 
In general, we find that the more forms of support to learn, the longer it may take for the online algorithms to ``converge.'' For various simulations, KNN is more consistent in learning a decision support policy efficiently. 


\section{Additional User Study Details and Results} \label{appdx:user_study}

\begin{figure}[htb]
    \centering
    \begin{minipage}{0.3\textwidth}
        \centering
        \includegraphics[width=0.95\textwidth]{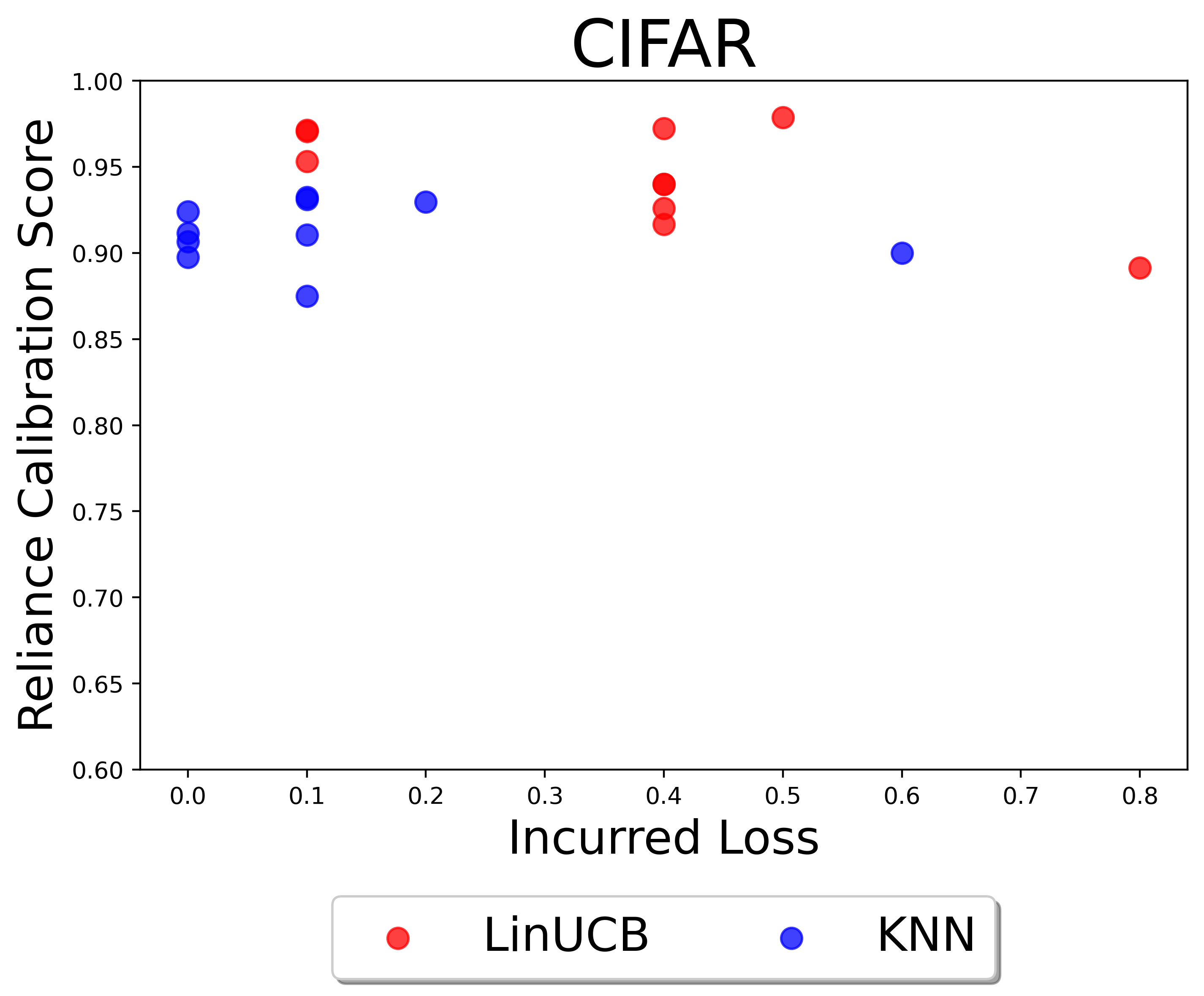} 
    \end{minipage}
    \begin{minipage}{0.3\textwidth}
        \centering
        \includegraphics[width=0.95\textwidth]{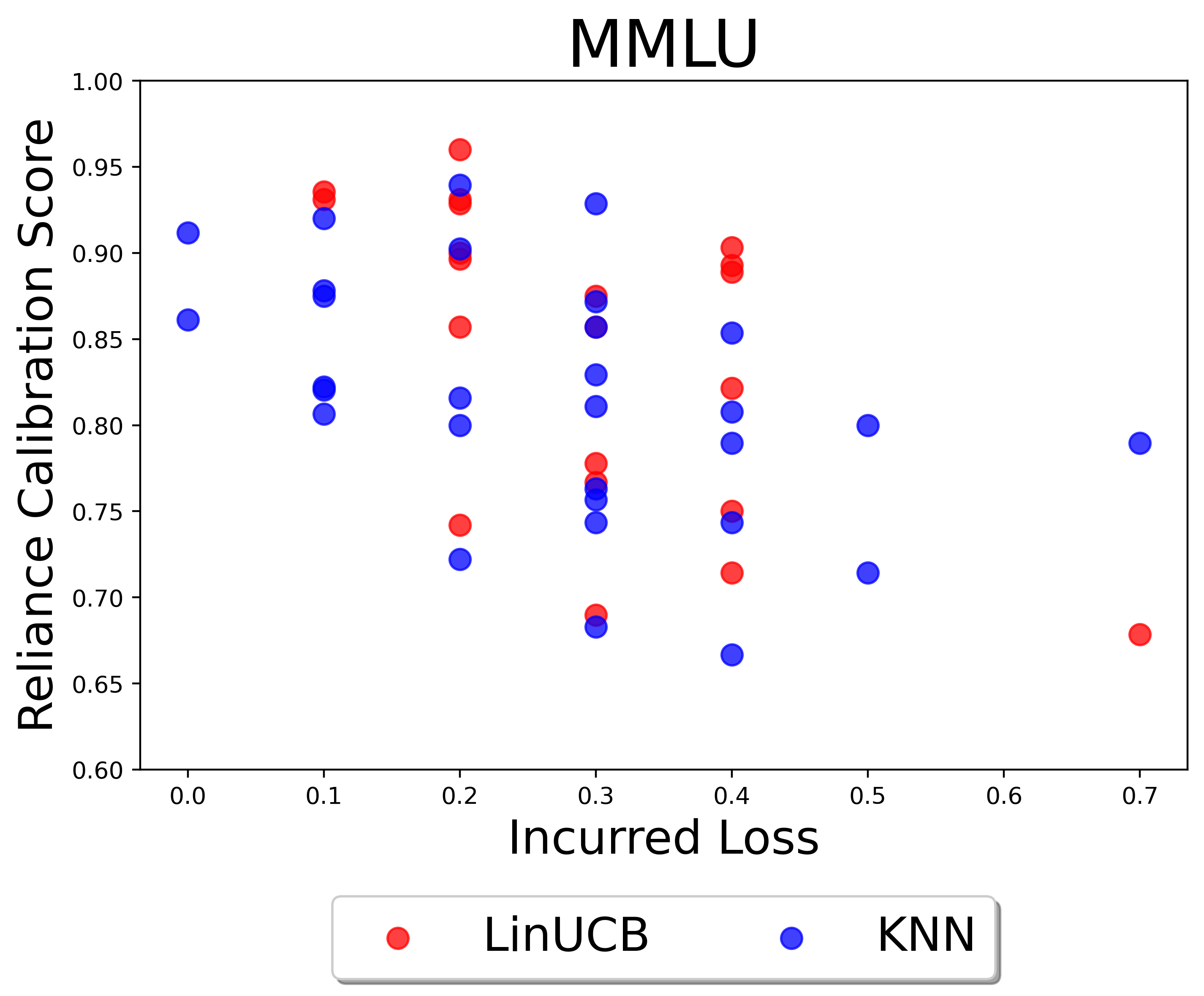} 
    \end{minipage}
    \caption{Relationship between a participant's sensibility of reliance (measured as the proportion of times they correctly agreed or disagreed with the form of support's prediction) and the loss incurred by the participant's learned policy. In the MMLU task, Modiste performs best for participants who sensibly rely on the provided support. Reliance is computed over all trials (100 for CIFAR-$3A$, 60 for MMLU-$2A$), and loss is averaged over the final 10 timesteps.
    }
        \label{fig:hse-reliance}
\end{figure}

\begin{figure*}[htb]
     \centering
     \includegraphics[width=0.7\linewidth]{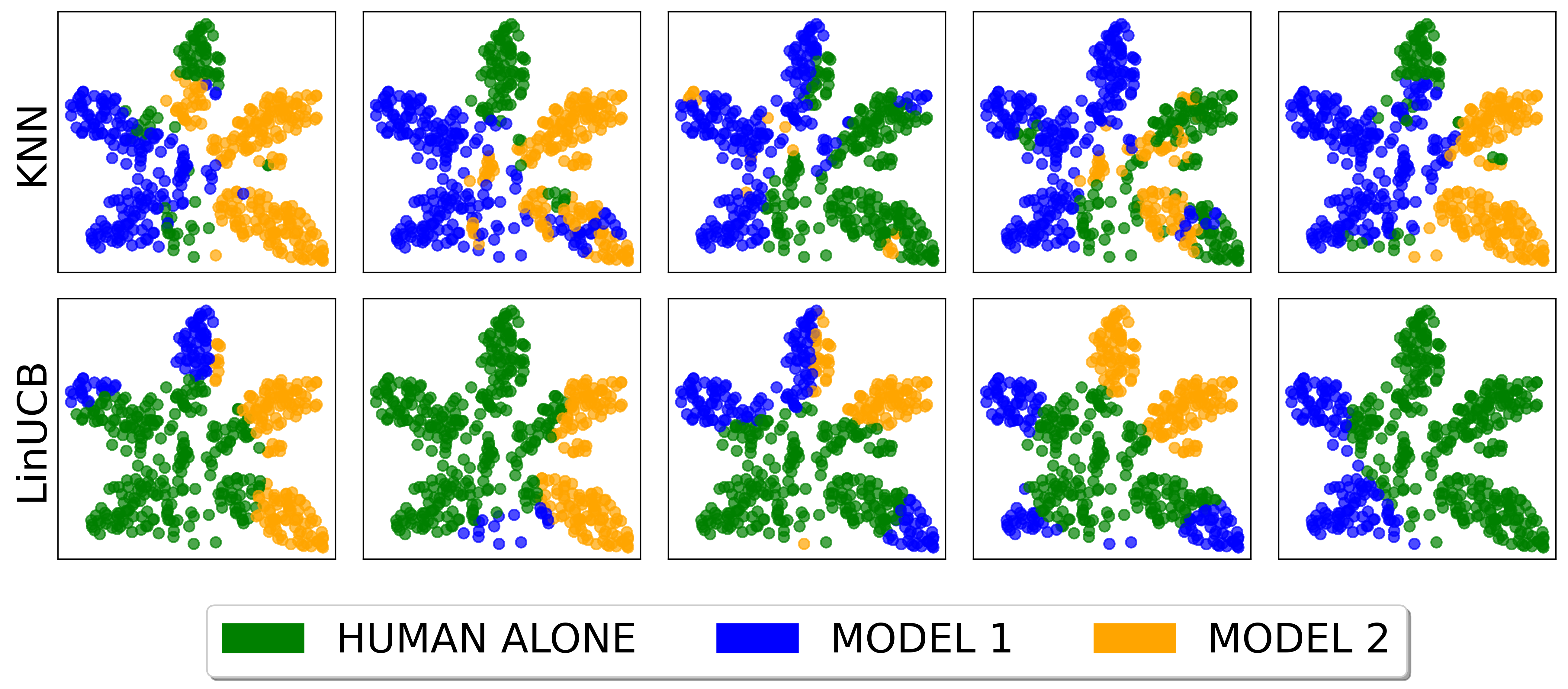}
     \caption{Snapshots of the recommended forms of support learned via \texttt{Modiste} for all participants in the CIFAR task. Policies learned using KNN get closer to optimal; see Figure \ref{fig:hse-cifar-topology}.}

    \label{fig:hse-topology-more-cifar}
\end{figure*}

\subsubsection{Algorithm Parameters}

For KNN, we take $K = 8$, $W = 25$, and $\epsilon=0.1$; for both algorithms, we let $\alpha = 1$. 
\footnote{Due to a server-side glitch, 6 of the CIFAR 125 recruited participants received incorrect feedback on $\leq 2\%$ of trials.} We use the same parameters for both tasks.



\subsection{Further Results}

\textbf{Visualizing Learned Policies.} We provide additional details on how the policy generalization snapshots in Figures \ref{fig:hse-cifar-topology} and \ref{fig:hse-mmlu-topology} are constructed. We save out the parameters learned by each algorithm while a participant interacts with \texttt{Modiste}. After the 60 or 100 trials (depending on which task they participated in), we load in the final state of parameters. We sample the embeddings of \textit{unseen points} from the respective task dataset (as described in Appendix~\ref{appdx:exp_details}) and pass these through the respective algorithm (LinUCB or KNN, depending on which variant the participant was assigned to) -- yielding a recommended form of support for said embedding. We color the latent by the form of support. We draw randomly 250 such unseen examples for CIFAR, and 100 for MMLU, when computing the policy recommendations.


We include snapshots for participants, across both tasks, in the main text. 
We depict all CIFAR examples and a random sampling of 7 of the 10 participants per variant for MMLU.


\subsubsection*{Participant Comments Hint at Mental Models of AI}

In a post-experiment survey where we permitted participants to leave general comments, we observed that several participants in our MMLU task noted that their responses were biased by the AI. We include a sampling of comments provided by participants which we found particularly revealing. We believe that these responses motivate the need for further study of how users' mental models of AI systems inform their decision-making performance, particularly in light of the rapidly advancing and public-facing foundation models~\citep{foundationModels}. 

\begin{itemize}
    \item \texttt{``I thought the AI answers went against my previous notions about what an AI would be proficient at answering. I thought an AI would excel at anything where there is a set answer such as math, computer science coding question, and biology textbook definitions. Then I expected it to absoutly fail at political, and historical questions. Since they can be more opinionated and not have a definitive answer. However, the AI struggled with math and excelled in the other topics. So I was actively not choosing to agree with the AI on any math where I could also easily check its work and was heavily relying on it for everything else [...] I knew nothing about any question on computer science, so I mainly relied on AI for answers on that topic.''}
    \item \texttt{``AI seemed to me more helpful in topics I didn't know, but sometime I trusted it too much, instead of thinking the answer through''}
    \item \texttt{``The AI had me second guessing myself sometimes so it affected my answers occasionally.''}
    \item \texttt{``Anything highlighted with the AI I was confident that would be the right answer.''}
\end{itemize}

\section{Further Discussion}

Our human subject experiments validate trends observed in the computational experiments, demonstrating that \texttt{Modiste} can be used to learn decision support policies online. 
We find that learning personalized policies leads to performance benefits when encountering users with ``varying'' expertise profiles, and recovers fixed policies for participants with ``strictly better'' profiles. 
Future work modeling human idiosyncrasies~\citep{steyvers2022three} explicitly in synthetic decision-makers may bolster the effectiveness of personalization.

Participants' reliance is one such factor.  We define a metric for ``reliance sensibility,'' as the proportion of trials where the participant agreed with the support when correct and responded differently when the support was incorrect, out of all trials where the participant received support.
The higher the proportion, the more ``sensible'' a participant's degree of reliance on the provided support is, relative to what would be most beneficial for decision outcomes.\footnote{Note that we cannot directly deduce whether, or to what degree, an individual participant relied on the form of support when presented - since we do not have access to what an individual would have said without support; alternative reliance inference schemes like \citep{tejeda2022ai} could be considered in future work.} 
We plot the performance of the learned policy (i.e., incurred loss) against a participant's reliance sensibility in Figure \ref{fig:hse-reliance}.
In CIFAR-$3A$, participants largely had calibrated reliance because, by design, there was a signal for when to rely on each form of support.
However, in the MMLU setting, we observe a variety of reliance behavior spanning over- and under-reliance, as we would expect in-the-wild~\citep{zerilli2022transparency}. 
We also find a negative correlation between expected loss and reliance (Pearson $r$ correlation $= -0.47$ and $-0.59$ for KNN and LinUCB, respectively), indicating that \texttt{Modiste} was less effective for participants who did not appropriately rely on support. 
At present, \texttt{Modiste} only considers expertise explicitly when personalizing policies, yet these data underscore the importance of understanding when individual decision-makers may over-rely on potentially fallible support~\citep{bussone2015role}. 
We explore practical considerations of personalized policies next.

An important concern of \texttt{Modiste} stems from our encouragement for individuals to \textit{disuse} decision support. 
In particular, as machine learning models are increasingly used as a form of decision-support~\cite{lai2021towards}, our approach to learning a decision support policy promotes the disuse of models in favor of human judgment and agency. 
Automating access to model outputs may disadvantage individuals whose behavior (i.e., observed decisions) is misaligned with their values and may be used by organizations to prevent access to model outputs in a ``big brother'' sense~\cite{zuboff2023age}.
We acknowledge that future work may need to justify to decision-makers when they are barred from accessing the model. 
This may lead to important intersections with the legal and policy communities. 
Instead of fully restricting access, less strict forms of disuse can be used to friction the decision-maker's experience with a model, while having the same effect as barring access: examples include hiding the model output, which can be clicked on to appear~\cite{buccinca2021trust}, or displaying model confidence only for uncertain inputs~\cite{antoran2020getting}.

Additionally, since personalization inherently requires the use of participatory data~\cite{james2023participatory}, there will likely need to be explicit consent for the data to be used in learning decision support policies. 
Though our personalized policies make no assumptions about access to data from individuals a priori when learning a decision support policy, we use data collected from a like population, but we imagine that similar approaches can be used to design policies with offline user data~\cite{charusaie2022sample,buccinca2024towards}. 
There may be unintended, negative impacts of collecting such data as organizations can use this data to implement nefarious decision support policies, whereby members are manipulated into using decision support, or to track user behavior. 
While we hope that personalized decision support policies encourage the thoughtful use of decision support, we caution against the ways that adverse actors can leverage these tools to privilege a select few, who in turn reap the benefits of personalized access to decision support tools~\citep{gabriel2024ethics}.

Finally, the settings in which our tool, \texttt{Modiste}, was evaluated targeted English-language-speaking crowdworkers. The ecological validity~\cite{zerilli2022transparency} of our results may not hold for other groups, whose experiences, values, and preferences with respect to the use of machine learning models in decision-making may differ. 
While we did receive requisite approval to learn personalized policies on human subjects from our institution's IRB, we hope that future work can cast a wider net for running experiments on groups of differing positionality~\citep{kirk2024prism}.
Relatedly, our human subject experiments focused on two tasks (CIFAR and MMLU), which are relatively low-stakes decisions.
We encourage future work to study similar problem formulations for personalized support in higher-stakes settings.
We sincerely hope that decision support policies become a pervasive mechanism for empowering individuals to take control when models are provided to them.

\end{document}